\documentclass[usenames,dvipsnames,nohyperref]{article}

\usepackage{gatingpreamble}
\usepackage[T1]{fontenc}
\usepackage{etoolbox}

\usepackage[normalem]{ulem}
\def\newdisplayskip{7pt}
\def\newdisplayshortskip{2pt}
\expandafter\def\expandafter\normalsize\expandafter{%
    \normalsize
    \setlength\abovedisplayskip{\newdisplayskip}
    \setlength\belowdisplayskip{\newdisplayskip}
    \setlength\abovedisplayshortskip{\newdisplayshortskip}
    \setlength\belowdisplayshortskip{\newdisplayshortskip}
}
\usepackage{enumitem}%
\setlist[description]{noitemsep, topsep=0pt}

\newcommand{\graph}{\Gamma} %
\newcommand{\edges}{\mathbf{E}} %
\newcommand{\nodes}{\mathbf{V}} %
\renewcommand{\v}{\nodes}
\newcommand{\e}{\edges}
\newcommand\source[1][]{%
  \ifstrempty{#1}
  {s}
  {s(#1)}}
\newcommand\target[1][]{%
  \ifstrempty{#1}
  {t}
  {t(#1)}}
\newcommand{\innodes}{\textrm{In}} %
\newcommand{\outnodes}{\textrm{Out}} %

\renewcommand{\dim}[1]{|#1|}
\newcommand{\unique}{}

\begin{document}

\twocolumn[
  \icmltitle{The Neural Race Reduction: Dynamics of Abstraction in Gated Networks}

  \icmlsetsymbol{equal}{*}

  \begin{icmlauthorlist}
    \icmlauthor{Andrew M. Saxe*}{ucl,meta,cifar}
    \icmlauthor{Shagun Sodhani*}{meta}
    \icmlauthor{Sam Lewallen}{ucl}
  \end{icmlauthorlist}

  \icmlaffiliation{ucl}{Gatsby Computational Neuroscience Unit \& Sainsbury Wellcome Centre, UCL}
  \icmlaffiliation{meta}{FAIR, Meta AI}
\icmlaffiliation{cifar}{CIFAR Azrieli Global Scholar, CIFAR}

  \icmlcorrespondingauthor{Andrew Saxe}{a.saxe@ucl.ac.uk}

  \vskip 0.3in
]

\printAffiliationsAndNotice{\icmlEqualContribution} %

\begin{abstract}

Our theoretical understanding of deep learning has not kept pace with its empirical success. While network architecture is known to be critical, we do not yet understand its effect on learned representations and network behavior, or how this architecture should reflect task structure.In this work, we begin to address this gap by introducing the Gated Deep Linear Network framework that schematizes how pathways of information flow impact learning dynamics within an architecture. Crucially, because of the gating, these networks can compute nonlinear functions of their input. We derive an exact reduction and, for certain cases, exact solutions to the dynamics of learning. Our analysis demonstrates that the learning dynamics in structured networks can be conceptualized as a neural race with an implicit bias towards shared representations, which then govern the model's ability to systematically generalize, multi-task, and transfer. We validate our key insights on naturalistic datasets and with relaxed assumptions. Taken together, our work gives rise to general hypotheses relating neural architecture to learning and provides a mathematical approach towards understanding the design of more complex architectures and the role of modularity and compositionality in solving real-world problems. The code and results are available at \url{https://www.saxelab.org/gated-dln}.

\end{abstract}

\section{Introduction}
\label{introduction}

\begin{figure}
  \begin{center}
    \includegraphics[width=.4\textwidth]{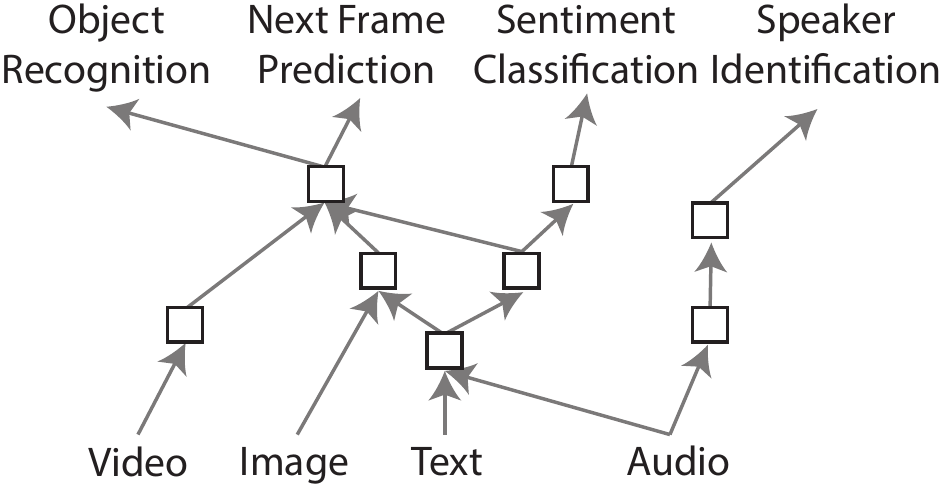}
  \end{center}
  \vspace{-5mm}
  \caption{A multi-modal network composed of simple modules shared across modalities and tasks. How do shared modules and pathways impact representation learning and generalization?}
    \label{fig:motivation}
    \vspace{-7mm}
\end{figure}

While neural networks have led to numerous impressive breakthroughs~\cite{resnet, transformer, amodei2016deep, baevski2020wav2vec, nature_dqn, alpha_zero}, our theoretical understanding of these models has not advanced at the same pace. Although we have gained some insight into general principles of network optimization~\cite{patel2015probabilistic,Carleo2019,bahri_statistical_2020,arora2020theory,roberts2021principles}, we have a limited understanding of how the specific choice of architecture---that is, the mesoscale pattern of connectivity between hidden layers (Fig.~\ref{fig:motivation})---affects a network's behavior~\cite{zagoruyko2016wide, raghu2017expressive,Chizat2018,saxe_mathematical_2019,tian2019luck}. For example, when training networks on the ImageNet dataset, wide networks perform slightly better on classes reflecting scenes, whereas deep networks are slightly more accurate on classes related to consumer goods~\cite{nguyen2020wide}. Understanding the reasons behind these behaviors may lead to more systematic techniques for designing neural networks.

Here, we address this gap by introducing and analyzing the Gated Deep Linear Network (GDLN) framework, which illuminates the interplay between data statistics, architecture, and learning dynamics. We ground the framework in the Deep Linear Network (DLN) setting as it is amenable to mathematical analysis, and previous works~\citep{Saxe2014,lampinen2018analytic,arora2018optimization,saxe_mathematical_2019} have observed that DLNs exhibit several nonlinear phenomena that are observed in deep neural networks. Because of the gating mechanism, however, GDLNs can compute nonlinear functions of their input, making them more expressive than standard deep linear networks.

Our main contributions are: (i) We introduce the GDLN framework (\cref{sec:gdln_framework}), which schematizes how pathways of information flow impact learning dynamics within an architecture. (ii) We derive an exact reduction and, for certain cases, exact solutions to the dynamics of learning (\cref{sec:gradient_flow_dynamics}). (iii) Our analyses reveal the dynamics of learning in structured networks can be conceptualized as a neural race with an implicit bias towards shared representations (\cref{sec:neuralracered}). (iv) We validate our findings on naturalistic datasets, with some relaxed assumptions (\cref{sec:experiment}).

\section{Gated Deep Linear Network Framework}
\label{sec:gdln_framework}

A fundamental principle of neural network design is that powerful networks can be composed out of simple modules~\cite{goodfellow2016deep, lecun2015deep}, and a key intuition is that a network's compositional architecture should resonate in some way with the task to be performed. For instance, a multimodal network might process each modality independently before merging these streams for further processing \cite{ngiam2011multimodal,girdhar2022omnivore}, and a multi-tasking NLP model might process its input through a shared encoder before it splits into task-specific pathways~\cite{collobert2011natural,liu2019multi,standley2020tasks}. Frequently, a network's compositional structure can be conceptualized as an ``architecture graph'' $\graph$, decorated by network modules, additional interactions, and learning mechanisms.

\begin{figure}
  \begin{center}
    \includegraphics[width=.4\textwidth]{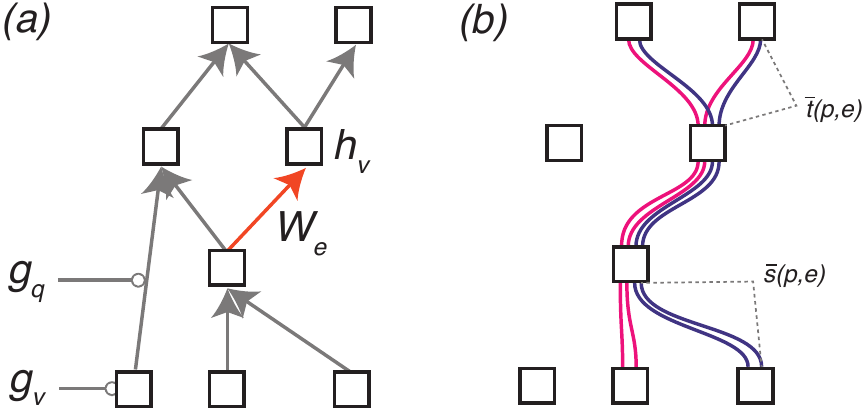}
  \end{center}
  \vspace{-3mm}
  \caption{Formalism and notation. (a) The Gated Deep Linear Network applies gating variables to nodes and edges in an otherwise linear network. (b) The gradient for an edge (here using the orange edge in Panel (a) as an example) can be written in terms of paths through that edge (colored lines). Each path is broken into the component antecedent $\bar s(p,e)$ and subsequent $\bar t(p,e)$ to the edge.}
    \label{fig:formalism}
    \vspace{-7mm}
\end{figure}

Here we introduce a class of networks, \textit{Gated Deep Linear Networks} (GDLNs), depicted in Fig.~\ref{fig:formalism}a, for which we can analytically study the effect of the architecture graph on learning and generalization. A GDLN is defined as follows. Let $\graph$ denote a directed graph with nodes $\nodes$ and edges $\edges$, with the structure of $\graph$ encoded by functions $\source, \target : \edges \to \nodes$ mapping an edge to its source and target node, respectively. For each $v\in \v$, let $\dim{v}\in \N$ denote the number of neurons assign to that node, and let $h_v \in \R^{\dim{v}}$ denote neural activations for the corresponding network layer. For each edge $e \in \e$, let $W_e$ denote a $\dim{\target[e]}  \times \dim{\source[e]}$ weight matrix assigned to $e$. An \textit{input node} of $\graph$ is a node with only outgoing edges, and an \textit{output} node is a node with only incoming edges. Let $\innodes(\graph), \outnodes(\graph) \subset V$ denote the sets of input nodes and output nodes of $\graph$, respectively.

The GDLN associated with $\graph$ computes a function as follows: An input example specifies values $x_v\in \R^{\dim{v}}$ for all input nodes $v\in \innodes(\graph)$, and the input nodes are fixed to their values $h_v=x_v$ for $v\in \innodes(\graph)$. Then, activation propagates to subsequent layers according to $h_v=g_v\sum_{q \in \edges:t(q)=v} g_qW_qh_{s(q)}$ where $g_v$ is the \textit{node gate} and $g_q$ is the \textit{edge gate}. That is, activity propagates through the network as in standard neural networks, but modulated by \textit{gating variables} that can act at the node and edge level.

In essence, these gating variables enable nonlinear computation from input to output, and can be interpreted in several ways: The node level gating can be viewed as an approximate reduction of ReLU dynamics as a ReLU neuron's activity can be written as $\max(0,wx)=\textrm{step}(wx)wx$. The gating variables can also be viewed as context-dependent control signals that modulate processing in the network. We discuss the extensive connections between the Gated Deep Linear Network and other  approaches in Appendix \ref{apx:related_work}.

In order to keep the analysis mathematically tractable, we assume that the gating variables are simply specified directly for each input that will be processed and we consider them to be a part of the dataset.

\section{Gradient Flow Dynamics and Reduction}
\label{sec:gradient_flow_dynamics}
Having described the design of the network, we now exploit its special properties to understand learning dynamics and their relationship to network structure.
In particular, we consider training all weights in the network to minimize the $L_2$ loss averaged over a dataset, \begin{equation}
  \mathcal{L}(\{W\})=\left\langle \frac{1}{2} \sum_{v\in \outnodes(\graph) } ||y_v - h_v ||_2^2 \right\rangle_{x,y,g}
\end{equation} where $y_v\in \R^{\dim{v}}$ for $v\in \outnodes(\graph)$ are the target outputs for the output layers in the network, and $\langle \cdot \rangle_{x,y,g}$ denotes the average over the training dataset and gating structures. The weights in the network can be updated using gradient descent. A key virtue of the GDLN formalism is that the gradient flow equations can be compactly expressed in terms of the \unique paths through the network.

We first lay out our notation, as illustrated in Fig.~\ref{fig:formalism}b for an example network. A path $p$ is a sequence of edges that joins a sequence of nodes in the network graph $\graph$. Let $\mathcal P(e)$ be the set of all \unique paths from any input node to any output node that pass through edge $e$. Let $\mathcal T(v)$ be the set of all \unique paths terminating at node $v$. We denote the source and the target node of the path $p$ as $s(p)$ and $t(p)$, respectively. We denote the component of path $p$ that is subsequent to edge $e$ (i.e., the path whose source node is the target of $e$, and that otherwise follows $p$) as $\bar t(p,e)$ (for the `target' path of $e$). Similarly, we denote the component of path $p$ that precedes edge $e$ (i.e., the path whose target node is the source of $e$, and that otherwise follows $p$) as $\bar s(p,e)$ (for the `source' path of $e$). Overloading the notation, we will write $W_p$ where $p$ is a path to indicate the ordered product of all weights along the path $p$, with the target of $p$ on the left and the source of $p$ on the right. Similarly, we write $g_p$ where $p$ is a path to denote the product of the (node and edge) gating variables along the path.

With this notation, the gradient flow equations can be shown to be (full derivation in Appendix \ref{apx:gradientflow}),
\begin{eqnarray}
  \tau \frac{d}{dt}W_e  &= & -\frac{\partial \mathcal L(\{W\})}{\partial W_e} \quad \forall e \in E, \\
  &= &  \sum_{p\in \mathcal P(e)} W_{\bar t(p,e)}^T\mathcal{E}(p)W^T_{\bar s(p,e)} \label{eq:grad_flow}
\end{eqnarray}
where the error term for path $p$ is
\begin{eqnarray}
  \mathcal{E}(p) = \Sigma^{yx}(p) - \sum_{j \in \mathcal T(t(p))} W_j\Sigma^{x}(j,p). \label{eq:error_matrices}
\end{eqnarray}
Here the dataset statistics which drive learning are collected in the correlation matrices
\begin{eqnarray}
  \Sigma^{yx}(p)&=& \left\langle g_p y_{t(p)} x_{s(p)}^T\right\rangle_{y,x,g} \label{eq:datastatsxy} \\
  \Sigma^x(j,p)&=&\left\langle g_jx_{s(j)}x_{s(p)}^Tg_p\right\rangle_{y,x,g} \label{eq:datastatsxx}
\end{eqnarray}
where $j$ and $p$ index two paths. Hence if there are $N$ \unique paths through the graph from input nodes to output nodes, there are potentially $N$ distinct input-output correlation matrices and $N^2$ distinct input correlation matrices that are relevant to the dynamics. Remarkably, no other statistics of the dataset are considered by the gradient descent dynamics.

Notably, these correlation matrices depend not just on the dataset statistics ($x$ and $y$), but also on the gating structure $g$. The possible gating structures are limited by the architecture. In this way, the architecture of the network influences its learning dynamics.

In essence, the core simplification enabled by the GDLN formalism is that the gating variables $g$ appear only in these data correlation matrices. They do not appear elsewhere in
Eqns.~\eqref{eq:grad_flow}-\eqref{eq:error_matrices}, which otherwise resemble the gradient flow for a deep linear network \cite{Saxe2014,saxe_mathematical_2019}.
The effect of the nonlinear gating can thus be viewed as constructing pathway-dependent dataset statistics that are fed to deep linear subnetworks (\unique pathways).

\begin{figure}
  \begin{center}
    \includegraphics[width=.5\textwidth]{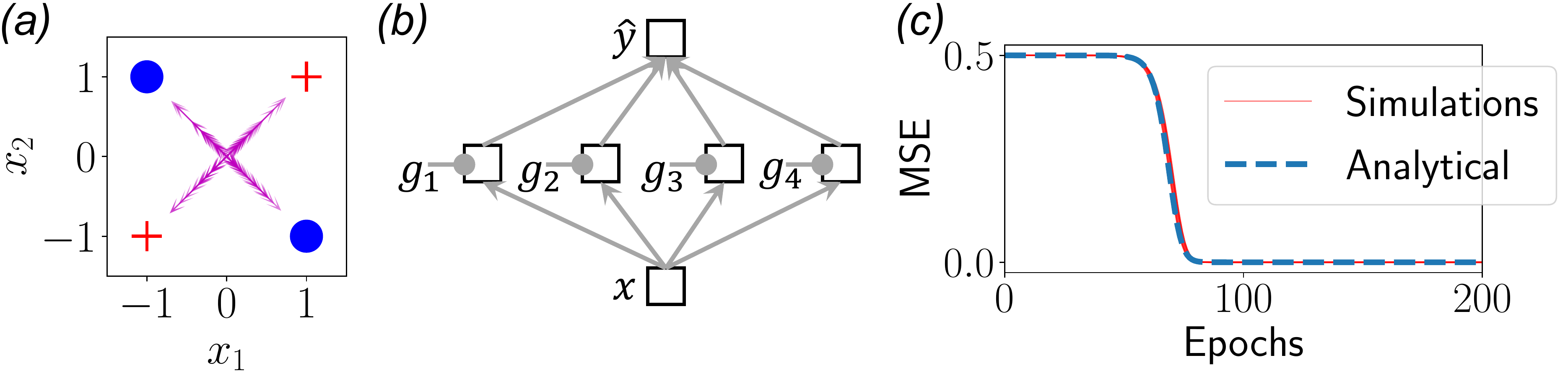}
  \end{center}
  \vspace{-3mm}
  \caption{XoR solution dynamics. (a) The XoR task with positive (red) and negative (blue) examples. Input-to-hidden weights from ReLU simulations (magenta) reveal four functional cell types. (b) GDLN with four paths, each active on one example. (c) Simulations of ReLU dynamics from small weights (red, 10 repetitions) closely track analytical solutions in the GDLN. \textit{Parameters:} $N_h=128,\tau=5/2,\sigma_0=.0002.$}
    \label{fig:xor}
    \vspace{-9mm}
\end{figure}

As a simple example of the power of this framework relative to deep linear networks, consider the XoR task (Fig.~\ref{fig:xor}a), a canonical nonlinear task that cannot be solved by linear networks. By choosing the gating structures to activate a different pathway on each example (Fig.~\ref{fig:xor}b), the gated deep linear network can solve this task (Fig.~\ref{fig:xor}c blue). Crucially, its dynamics (analytically obtained in~\cref{sec:nonlinear_contextual_classification} based on the reduction in the following sections) closely approximates the dynamics of a standard ReLU network trained with backprop (Fig.~\ref{fig:xor}c red). This result demonstrates that the gated networks are more expressive than their non-gated counterpart, and that gated networks can provide insight into ReLU dynamics in certain settings. We note that so far, our analysis does not provide a mechanism to select the gating structure. We will return to this point in Section \ref{sec:neuralracered}, which provides a perspective on the gating structures likely to emerge in large networks.

\subsection{Exact reduction from decoupled initial conditions}

Our fundamental goal is to understand the dynamics of learning as a function of architecture and dataset statistics. In this section, we exploit the simplified form of the gradient flow equations to obtain an exact reduction of the dynamics. Our reduction builds on prior work in deep linear networks, and intuitively, shows that the dynamics of gating networks can be expressed succinctly in terms of
effective independent 1D networks that govern the singular value dynamics of each weight matrix in
the network. The reduced dynamics can be substantially more compact, as for instance, a weight
matrix of size $N\times M$ has $NM$ entries but only $\textrm{min}(M,N)$ singular values.

\begin{figure*}[ht]
  \begin{center}
    \includegraphics[width=.8\textwidth]{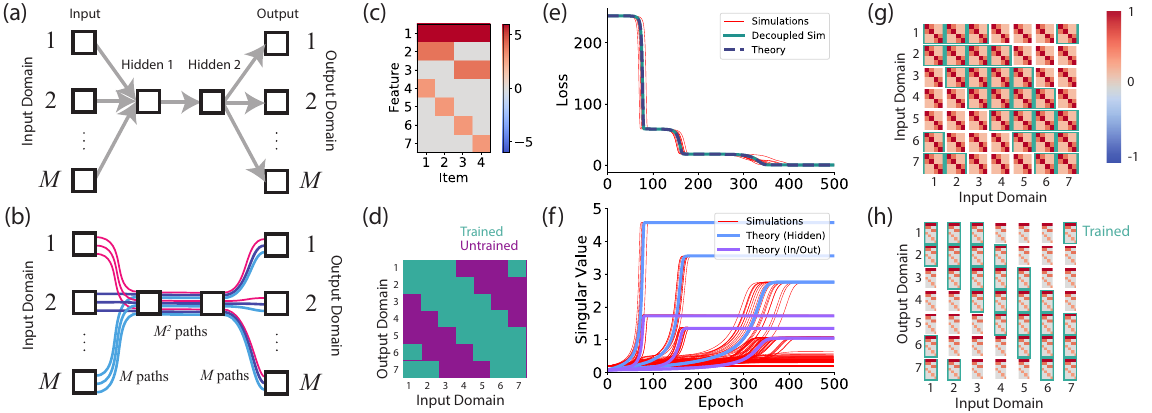}
  \end{center}
  \vspace{-4mm}
  \caption{Pathway network solution dynamics. (a) The network contains $M$
    different input domains (each consisting of a bank of neurons), $M$ different
    output domains, and two hidden layers. The task is to learn a mapping from
    each input domain to each output domain. The gating structure gates
    on one input and one output pathway. The hidden pathway is always
    on. (b) Gated network formalism. There are $M^2$ \unique pathways through the
    network from input to output. All $M^2$ flow through the hidden weight matrix, while
    only $M$ flow through each input or output weight matrix. This fact causes the hidden
    layer to learn faster. (c) Small example dataset with hierarchical structure.
    The task of the network is to produce the 7-dimensional output vector for each
    of four items. Inputs are random orthogonal vectors for each item. (d) Each
    input domain is trained with $K$ output domains (here $K=4$), such that some
    input-output routes are never seen in training. (e) Training loss dynamics for
    simulated networks from small random weights (red, 10 repetitions), simulated
    networks from decoupled initial conditions (green), and theoretical prediction
    from Eqn. \ref{eqn:scalar_routing_reduction} (blue). The theory matches the decoupled simulations
    exactly, and is a good approximation for small random weights. (f) The singular values of the hidden weight matrix (blue) are larger than those in input or output matrices by a factor $\sqrt{M}$. Theoretical predictions match simulations well, particularly for larger singular values. (g) Representational similarity (or kernel) matrix at the first hidden layer. Inputs from different domains are mapped to similar internal representations, revealing a shared representation even for input domains that are never trained with a common output. (h) Predicted output at the end of training. The network generalizes perfectly to input-output routes that were never seen during training.  \textit{Parameters:} $M=7,K=4,\lambda=.02,\sigma_0=.2,N_h=64.$}
    \label{fig:routing_network}
    \vspace{-5mm}
\end{figure*}

To accomplish this, we introduce a change of variables based on the singular value decomposition of the relevant dataset statistics. Suppose that the dataset correlation matrices are mutually diagonalizable, such that their singular value decompositions have the form
\begin{eqnarray}
  \Sigma^{yx}(p)&=& U_{t(p)}S(p) V_{s(p)}^T \label{eq:sig_yx}\\
  \Sigma^x(j,p)&=&V_{s(j)}D(j,p) V_{s(p)}^T
\end{eqnarray}
where the set of $U$ and $V$ matrices are orthogonal, and the set of $S$ and $D$ matrices are
diagonal. That is, there is a distinct orthogonal matrix $U_l$ for each output layer, a distinct
orthogonal matrix $V_l$ for each input layer, and diagonal matrices $S(p),D(p)$ for each \unique path through the network.

Then, following analyses in deep linear networks \cite{Saxe2014}, we consider the following change of variables. We rewrite the weight matrix on each edge
as
\begin{eqnarray}
  W_e(t) = R_{t(e)}B_e(t)R_{s(e)}^T \quad \forall e, \label{eq:change_of_var}
\end{eqnarray}
where the matrices $B_e(t)$ are the new dynamical variables, and the matrix $R_v$ associated to each node $v$ in the graph satisfies $R_v^TR_v=I$. Further, for output nodes $v$, we require $R_v=U_v$, the output singular vectors in the diagonalizability assumption. Similarly,
for input nodes, we require $R_v=V_v$.

Inserting \eqref{eq:sig_yx}-\eqref{eq:change_of_var} into \eqref{eq:grad_flow}-\eqref{eq:error_matrices} shows that the dynamics for $B_e$ decouple: if all $B_e(0)$ are initially diagonal, they will remain so under the dynamics (full derivation in Appendix \ref{apx:gradientflow}). For this decoupled initialization, the dynamics are
\begin{equation}
  \tau \frac{d}{dt}B_e  =   \sum_{p\in \mathcal P(e)} B_{p \setminus e}\left[S(p)  -  \sum_{j \in \mathcal T(t(p))} B_jD(j,p)  \right] \label{eq:decoulped_dyn}
\end{equation}
where $B_{p \setminus e}=B_{\bar t(p,e)}B_{\bar s(p,e)}$ is the product of all $B$ matrices on path $p$ after removing edge $e$ (see Appendix \ref{apx:gradientflow}).

In essence, this reduction removes competitive interactions between singular value modes, such
that the dynamics of the overall network can be described by summing together several ``1D
networks,'' one for each singular value. Intuitively, this reduction shows that learning dynamics depend on several factors.
\begin{description}
  \item[Input-output correlations] Other things being equal, a pathway learning from a dataset with larger input-output singular values will learn
faster. This fact is well known from prior work on deep linear networks \cite{Saxe2014}.
\item[Pathway counting] Other things being equal, a weight matrix corresponding to an edge that participates in many paths (such that the sum contains
many terms) will learn faster. This fact is less obvious, as it becomes relevant only if one moves
beyond simple feed-forward chains to study complex architectures and gating.
\end{description}
We now turn to examples that verify and illustrate the rich behavior and consequences of these dynamics.

\section{Applications and consequences}
\label{sec:applications}

To fix a specific scenario with rich opportunities for generalization, we consider a ``routing'' setting, as depicted in Fig.~\ref{fig:routing_network}a. In this setting, a network receives inputs from $M$ different input domains and produces outputs across $M$ different output domains. The goal is to learn to map inputs from a specific input domain to a specific output domain, with no negative-interference from other input-output domain pairs. There are thus $M^2$ possible tasks which can be performed, each corresponding to mapping one of the $M$ input domains to one of the $M$ output domains.

We assume that the target input-output mapping from the active input domain to the active output domain is the same for all pathways, and defined by a dataset with input correlations $\langle xx^T \rangle = VDV^T$ and input-output correlations $\langle yx^T\rangle = USV^T$. For the simulations in this section, we take the dataset to contain four examples, and the target output to be a 7-dimensional feature vector with hierarchical structure (Fig.~\ref{fig:routing_network}c), but note that the theory is more general.

To investigate the possibility of structured generalization, we consider a setting where only a subset of input-output pathways are trained. That is, each input domain is trained with only $K\leq M$ output domains, as depicted in  Fig.~\ref{fig:routing_network}d, such that some input-output pathways are never observed during training.

We consider solving this task with a two-hidden layer gated deep linear network depicted in Fig.~\ref{fig:routing_network}a. We emphasize that this task is fundamentally nonlinear, because inputs on irrelevant input domains must be ignored. We take the gating structure to gate off all first layer pathways except the relevant input domain, and to gate off all third layer pathways but the one to the relevant output domain. As shown in Fig.~\ref{fig:routing_network}b, this scheme results in $M^2$ pathways through this network that must be considered in the reduction. The resulting pathway correlations are simply given by the original dataset, scaled be the probability that each path is active (see Appendix \ref{apx:routing_network_reduction})
\begin{eqnarray}
  \Sigma^{yx}(p)&=&  \frac{1}{KM}USV^T\\
  \Sigma^x(j,p)&=&\begin{cases}
    \frac{1}{KM}VDV^T \quad \textrm{if}~j=p \\
    0 \quad \textrm{otherwise}
  \end{cases}.
\end{eqnarray}
Crucially, we have a simple ``pathway counting'' logic behind the reduction: the first and third layer weights are active in $K$ paths (all tasks originating from a given input domain or terminating at a given output domain, respectively), while second layer weights are active in $KM$ trained paths. This fact causes the second layer weights to learn more rapidly.

Assuming that weights start out roughly balanced in each first layer and third layer weigh matrix (a reasonable assumption when starting from small random weights), this yields the reduced dynamics (Appendix \ref{apx:routing_network_reduction})
\begin{eqnarray}
  \tau \frac{d}{dt}B_1 &=& \frac{1}{M} B_2B_1\left[S-B_2B_1^2D\right] \label{eqn:routing_net_reduct_eq1}\\
  \tau \frac{d}{dt}B_2 &=& B_1^2\left[S-B_2B_1^2D\right]\label{eqn:routing_net_reduct_eq2}
\end{eqnarray}
where $B_1$ describes the input and output pathway weights singular values, and $B_2$ describes the hidden layer weight singular values.

We note that the quantity $MB_1^2-B_2^2$ is conserved under the dynamics. Defining the constant $C=MB_1(0)^2-B_2(0)^2$, we can therefore write the dynamics as
\begin{equation}
    \tau \frac{d}{dt}B_2 = \frac{1}{M}(B_2^2+C)\left[S-\frac{1}{M}B_2(B_2^2+C)D\right]. \label{eqn:scalar_routing_reduction}
\end{equation}
Remarkably, this equation reveals that the dynamics of this potentially large, gated, multilayer network with arbitrary numbers of hidden neurons can be reduced to a single scalar for each singular value in the dataset. Each diagonal element of this equation provides a separable differential equation that may be integrated to give an exact formal solution.

Figure~\ref{fig:routing_network}e compares the training error dynamics predicted by Eqn.~\ref{eqn:scalar_routing_reduction} to full simulations starting from small random weights (i.e. scaling Xavier initialiation weights by 0.2), verifying that our reduction is a good description of dynamics starting from small random weights. Furthermore, as shown in Appendix \ref{apx:routing_network_reduction}, the hidden layer weight singular values change by a factor of $\sqrt{M}$ more than the input or output weights, as verified in Fig~\ref{fig:routing_network}f.

\begin{figure*}[h]
  \begin{center}
    \includegraphics[width=\textwidth]{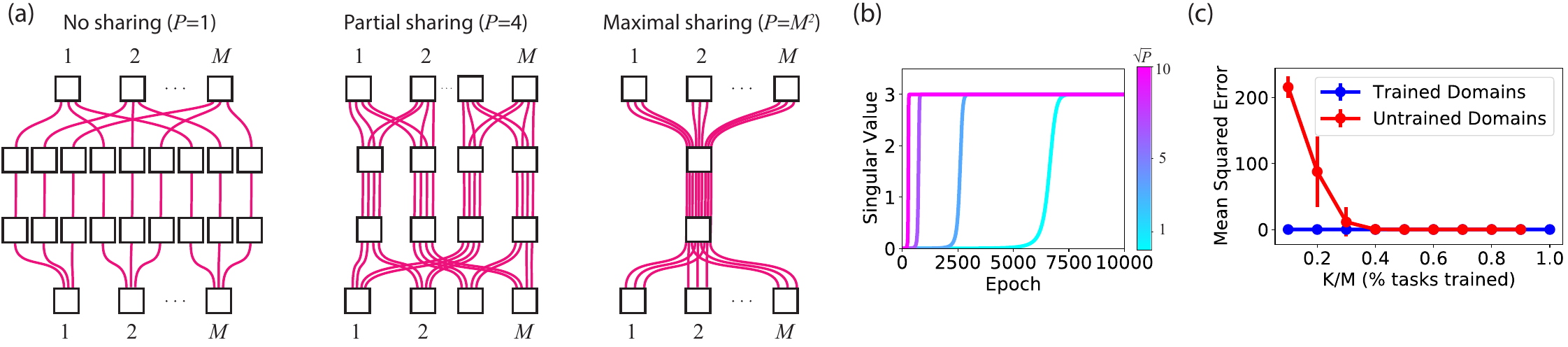}
  \end{center}
  \vspace{-4mm}
  \caption{Pathway race dynamics. (a) The same routing task can be solved using a variety of gating schemes that differ in their use of shared representations. Every input-output combination can be given a dedicated pathway such that $P=1$ tasks flow through each (left), groups of two input domains and two output domains can share a pathway such that $P=4$ tasks flow through each (middle), or all $P=M^2$ tasks can run through a shared representation (right). (b) Singular value dynamics as a function of the number of pathways $P$ flowing through the hidden layers. Networks that share more structure learn faster. Consequently, in a single network where subparts share structure to different degrees, the maximally shared dynamics dominate the race between pathways. (c) Error on trained (blue) and untrained (red) pathways as a function of the fraction of output domains $K$ trained with each input domain $M$. When few outputs are trained per input domain, the race dynamics do not strongly favor shared structure and so error on untrained domains is large. When $\sim 40\%$ of output domains are trained with each input, the shared structure solutions are sufficiently faster to reliably dominate the race and yield generalization to untrained domains. \textit{Parameters:} $M=10,\lambda=.05/K,\sigma_0=.2,N_h=64.$}
  \label{fig:pathway_race_red}
   \vspace{-5mm}
\end{figure*}

\subsection{Shared representations and generalization}
\label{sec:shared_representations_and_generalization}
With this description of the training dynamics of the network, we can then ask what representations emerge in the network over training. One way of interrogating the nature of representations in the network is to compute the representational similarity between different input examples from the same input domain, and across different input domains. Specifically, we compute the dot product between the neural activity in the first hidden layer in response to different inputs. As shown in Fig.~\ref{fig:routing_network}g, the pathway network learns a \textit{shared representation}, in which each individual example maps to the same representation regardless of what input domain it arrives on. That is, the gating and learning dynamics enable the network to learn a representation that is \textit{invariant} to input domain, and which is abstract in the sense that the representation contains no information about what input domain produced it. This shared representation supports zero-shot generalization to untrained input-output pathways, as shown in Fig.~\ref{fig:routing_network}h.

The intuition behind obtaining zero-shot generalization is as follows: Say that we are evaluating the network on a new input-output pair of domains. As long as the network has been trained on examples from the current input domain (in conjunction with any output domain), the network will map it to the shared representation. Similarly, as long as the network has been trained on examples from the current output domain, it will be able to map this shared representation to the output. In this way, training on a subset of $M^2$ tasks is enough to obtain strong generalization to all $M^2$ tasks.

In essence, this solution accomplishes a factorization of the problem into two interacting but distinct components: the gating variables represent what input domain links to what output domain, providing information about ``where'' signals should go; while the neural activity represents ``what'' task-relevant input was presented, regardless of where it came from or where it should be routed to. This factorization can permit generalization to untrained pathways provided the gating structure is configured appropriately.

\begin{figure*}
  \begin{center}
    \includegraphics[width=.8\textwidth]{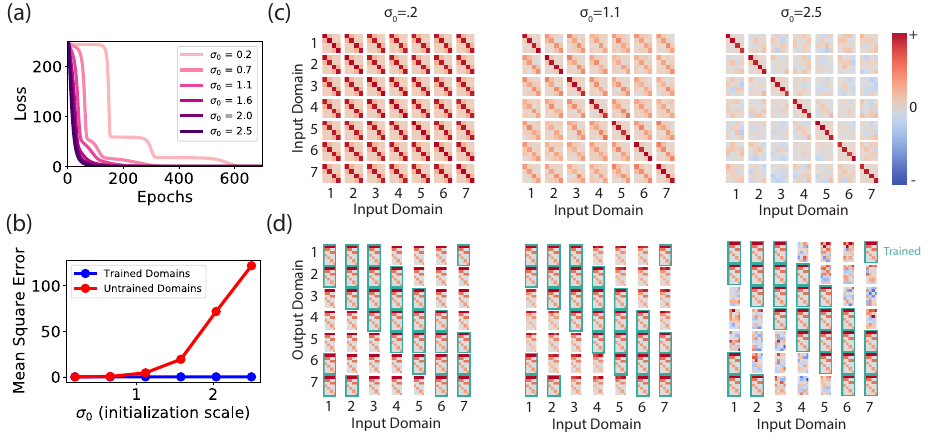}
  \end{center}
  \vspace{-5mm}
  \caption{Neural race vs. NTK regime: Initialization, shared representation formation, and zero-shot generalization. (a) Training loss for pathway networks with different initialization scales $\sigma_0$. Small $\sigma_0$ yields pathway race dynamics with stage-like drops through training. Large $\sigma_0$ yields NTK-like dynamics with rapid, exponential learning curves.   (b) Error for trained (blue) and untrained (red) input-output domain combinations as a function of initialization scale. While performance on trained domains is excellent for all scales, zero-shot generalization only emerges in the neural race regime. (c)  Representational similarity between inputs presented to different domains, for small (left column), medium (middle column) and large (right column) initialization scales. At small initialization scales, internal representations in the first hidden layer are similar even across domains, indicating one common shared representation. Large initialization scales place the network in the NTK regime where random initial connectivity persists throughout learning, yielding distinct high-dimensional random representations for each domain. (d) Network output for all input-output combinations for three initialization scales (labeled in panel c). Because networks in the NTK regime do not learn a shared representation for different input domains, they do not generalize to untrained pathways.  }
  \label{fig:pathway_vs_ntk}
  \vspace{-5mm}
\end{figure*}

\subsection{Neural race dynamics: Implicit bias toward shared representations}
\label{sec:neuralracered}

The reductions so far have assumed that the gating structure is specified \textit{a priori}, and furthermore, that different domains connect to the same singular value modes in the hidden layer. That is, the gating structure provides the opportunity for learning a shared representation, but this is not obligatory: different parts of the hidden representation could learn distinct pathways, despite all being gated on. Remarkably, the dynamics from small random weights track the trajectory predicted for maximally shared representation, suggesting that the full solution dynamics rapidly converge to the submanifold of decoupled weights. Why are shared representations favored under the dynamics?

To investigate this, we note that the same task and architecture typically permit several gating schemes and singular value mode connectivity patterns that each would obtain zero training error. As shown in Fig.~\ref{fig:pathway_race_red}a (left), for instance, the routing task could be solved with an alternative gating structure in which each input-output route receives a dedicated pathway that is gated on only when the task is to connect that specific input-output route. This gating scheme would still obtain zero training error, but does not yield any representation sharing. Other partial sharing schemes are possible; for instance, representations could be shared across groups of two input and two output domains (Fig.~\ref{fig:pathway_race_red}a, middle). What is the impact of these choices on learning dynamics and generalization?

Taking the case where all routes are trained ($K=M$) for simplicity, with no sharing, each pathway participates in just one of the $M^2$ total trained pathways, compared to the fully shared solution where the input and output layers participate in $M$ pathways and the hidden layer participates in $M^2$. From this, we can see that greater sharing leads to faster learning. In a combined network that produces its output using both shared and non-shared representations, the dynamics of each pathway will race each other to solve
the task; and hence the most-shared structure will dominate.

To see this quantitatively, we repeat a similar derivation to the preceding section for networks with varying degrees of pathway overlap. In particular, we parameterize the degree of pathway overlap with the parameter $P$ that counts the number of pathways flowing through a given hidden layer. The resulting reduction is
\begin{eqnarray}
  \tau \frac{d}{dt}B_1 &=& \frac{\sqrt{P}}{M^2} B_2B_1\left[S-B_2B_1^2D\right] \\
  \tau \frac{d}{dt}B_2 &=& \frac{P}{M^2} B_1^2\left[S-B_2B_1^2D\right],
\end{eqnarray}

which shows that the learning rate in all layers increases as $P$ increases. Solution dynamics for a range of degrees of sharing $P$ are plotted in Fig.~\ref{fig:pathway_race_red}b, which show that greater degrees of sharing reliably leads to faster singular value dynamics.

Hence, dynamics in GDLNs take the form of a \textit{pathway race}: when many gating schemes coexist in the same network, the ones that share the most structure--and hence learn the fastest--will come to dominate the solution. Therefore gradient flow dynamics in complex network architectures has an implicit bias toward extracting shared representations where possible. The strength of this bias increases as each input domain is trained with more output domains. As shown in Fig.~\ref{fig:pathway_race_red}c, shared representations begin to dominate reliably when roughly 40\% of input-output routes are trained, enabling generalization to unseen input-output routes.

\subsection{Impact of initialization}
\label{sec:initialiation}

The training and generalization dynamics of deep networks are known to depend on the weight initialization. Here we show that initial weight variance exerts a pronounced effect on the emergence of shared representations, and hence generalization abilities. As observed in a number of theoretical and empirical works, neural networks can operate in two different initialization regimes~\cite{Chizat2018,bahri_statistical_2020,flesch_orthogonal_2022}. Sufficiently wide networks initialized with large variance initializations enter the Neural Tangent Kernel regime, where training dynamics follow a simple linear dynamical system and error trajectories exhibit exponential approach to their asymptote \cite{jacot2018neural,Lee2019,arora_exact_2019}. Intuitively, in this regime, the initial strong random connectivity in the network provides sufficiently rich features to learn the task without substantially changing internal representations. In this setting, deep networks behave like kernel machines with a \textit{fixed} kernel (the neural tangent kernel). By contrast, networks initialized with sufficiently small variance initializations learn rich task-specific representations, and their dynamics as we have seen can be more complex \cite{mei_mean_2018,rotskoff_parameters_2018,sirignano_mean_2020,saxe_mathematical_2019}.

To show the effect of this transition in our setting, we train pathway networks starting from different random matrices with singular value $\sigma_0$. For a range of initialization scales, all networks converge to zero training error (Fig.~\ref{fig:pathway_vs_ntk}a). As expected, large initialization scales lead to NTK-like exponential dynamics, while small initialization scales lead to progressive stage-like drops in the error consistent with prior analyses of deep linear networks in the rich regime. Critically, initialization has a dramatic impact on generalization (Fig.~\ref{fig:pathway_vs_ntk}b), and only small initializations are capable of zero-shot generalization to untrained routes. To understand why, we visualize the representational similarity structure for several networks in Fig.~\ref{fig:pathway_vs_ntk}c, which shows a transition from shared to independent representations as networks move from the rich to the lazy regime. Finally, Fig.~\ref{fig:pathway_vs_ntk}d shows the breakdown in generalization in the NTK regime. Hence our neural race reduction can describe learning in the rich feature learning regime, with non-trivial generalization behavior.

\begin{figure*}[ht]
  \begin{center}        
    \includegraphics[width=\textwidth]{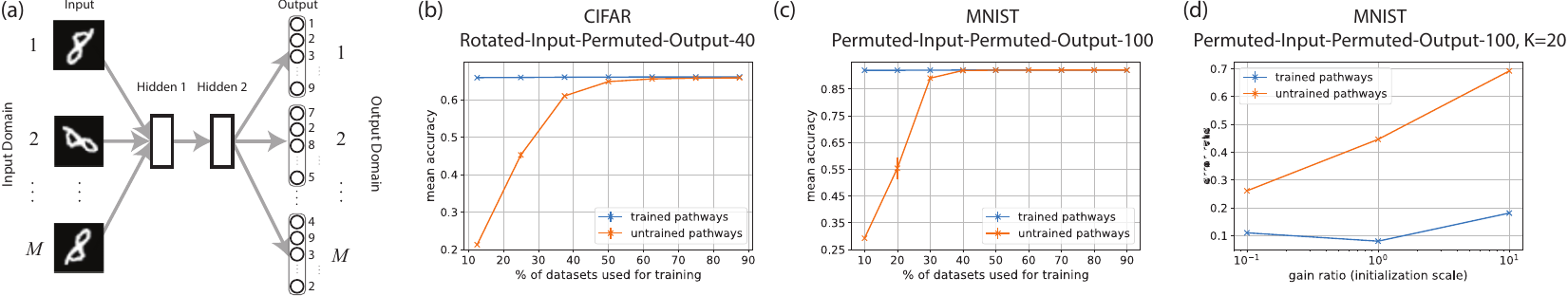}
  \end{center}
  \vspace{-5mm}
  \caption{Experimental results. (a) Each input domain receives inputs that have been subjected to one of $M$ input transformations. The target output for each output domain is also transformed by one of $M$ output transformations. Here the visualization uses rotations in the input and permutations in the output. Only a subset of input-output transformation pairs are seen during training. (b) Error for trained (blue) and untrained (orange) input-output domain pairs as a function of the percentage of trained pathways ($K/M$) on the CIFAR dataset with $M^2=1600$ total tasks. (c) Error on MNIST with $M^2=10^4$ total tasks. Training accuracy is always high while zero-shot transfer to untrained pathways becomes as good as the training performance when $\approx$40\% of pathways are trained. (d) Error as a function of initialization scale. While performance on trained domains is good for all scales, zero-shot generalization only emerges at small inits.} 
  \vspace{-5mm}
\label{fig:results_on_natural_images}
\end{figure*}

\section{Experiments}
\label{sec:experiment}

So far, we have described mathematical principles relating network architecture to learning dynamics and the nature of learned representations in GDLNs. We demonstrated how the network architecture affects generalization in pathway networks when applied to a simple toy dataset. In this section, we qualitatively validate our key findings on naturalistic datasets. Specifically, we test if GDLNs can exhibit strong zero-shot generalization performance on~\textit{untrained} input-output domain pairs (\cref{sec:shared_representations_and_generalization}) and if the link between initialization scale and zero-shot generalization (\cref{sec:initialiation}) holds when training on naturalistic datasets.

Several datasets and benchmarks have been proposed to evaluate systematic generalization in neural networks~\cite{johnson2017clevr,bahdanau2018systematic, Sinha2019CLUTRRAD,lake2019compositional,ruis2020benchmark,Sinha2020EvaluatingLG}. However, these naturalistic datasets generally require the network to learn multiple capacities (like spatial reasoning, logical induction, etc.) and come with a fixed (and often unclear) extent of training in different domains. To develop a setting that remains close to the theory, we create new datasets by \textit{composing transformations} on popular vision datasets. Having fine-grained control over the dataset generation mechanism enables us to understand the effect of parameters like the number of input/output domains.

Briefly, as depicted in~\cref{fig:results_on_natural_images}a, starting from a base dataset with $n$ inputs and outputs $\{(x^\mu,y^\mu)\},\mu=1,\cdots,n$, we generate new tasks by applying one of $M$ input transformations $f^{\textrm{in}}_i,i=1,\cdots,M$ and one of $M$ output transformations $f^{\textrm{out}}_j,j=1,\cdots,M$ (full details deferred to Appendix \ref{apx:experiments} due to space constraints). The task from input domain $i$ to output domain $j$ thus has samples $\{(f^{\textrm{in}}_i(x^\mu),f^{\textrm{out}}_j(y^\mu))\}$. We use the MNIST ~\citep{deng2012mnist} and CIFAR-10 datasets~\citep{krizhevsky2009learning} as base datasets, and rotations and permutations as transformations. Relative to our previous experiments, these datasets add real data correlations and distinct transformations on each domain that make finding a shared representation challenging.

\subsection{Results}
\label{sec:experiment_results}

In~\cref{fig:results_on_natural_images}, we evaluate the zero-shot generalization performance of gated deep linear networks on ~\textit{untrained} pathways and study the effect of initialization on their performance. Full model and training details are given in Appendix \ref{apx:experiments}. Fig.~\ref{fig:results_on_natural_images}(b,c) shows mean accuracy, over trained and untrained pathways, as a function of the fraction of datasets that the model used for training, for CIFAR ($M=40$) and MNIST ($M=100$) respectively. Training accuracy is always high while the zero-shot transfer to untrained pathways becomes near-perfect when $\approx$40\% of pathways are trained. In Fig.~\ref{fig:results_on_natural_images}d, we report the error for trained (blue) and untrained (orange) input-output domain combinations as a function of initialization scale (gain ratio). While performance on trained domains is good for all scales, zero-shot generalization only emerges for smaller scales, as in the neural race regime. These observations validate our findings (from~\cref{sec:applications}) on naturalistic datasets.

\section{Related Work}
\label{apx:related_work}
Our work is closely related to several areas in machine learning: Deep Linear Networks, the study of dynamics of learning in neural networks, and modular neural networks.

\textbf{Deep Linear Networks}:~\citet{Baldi1989,Fukumizu1998,Saxe2014,arora2018optimization,lampinen2018analytic,saxe_mathematical_2019} showed that deep linear networks exhibit several nonlinear phenomena that are observed in deep neural networks and proposed studying the dynamics of deep linear networks as a surrogate for understanding the dynamics in the deep neural networks. \citep{Baldi1989} described the loss landscape, while \citep{Saxe2014} developed the theory of gradient descent learning in deep linear neural networks and provided exact solutions to the nonlinear dynamics. Motivated by their observation about the similarity in dynamics of linear and non-linear networks and the feasibility of analyzing the gradients in the linear network, we ground the proposed framework in the deep linear network setting.

Many works have studied the \textbf{dynamics of deep networks} under specific assumptions like linear separability of data~\cite{des2018learning}, deep ReLU networks~\cite{tian2019luck,straat2019line}, Tensor Switching Networks~\cite{NIPS2016_b1563a78} (generalization of ReLU to tensor-valued
hidden units), the Neural Tanget Kernel limit~\cite{jacot2018neural,fort2020deep}, the Mean Field limit~\cite{mei_mean_2018,rotskoff_parameters_2018,sirignano_mean_2020} and Ensemble Networks~\cite{fort2019deep} to name a few (see \cite{Carleo2019,bahri_statistical_2020,arora2020theory} for reviews). Similar to these works, we also focus on a specific subset of deep networks, gated deep linear networks, that captures a nonlinear relationship between a network’s input-output maps and its parameters, while being amenable to theoretical analysis.

Within the setup of deep linear networks, several works have focused on the analysis of convergence rate~\cite{Saxe2014,arora2018convergence, arora2018optimization,du2019width}, on understanding inductive biases like implicit regularization~\cite{laurent2018deep,ji2018gradient,gunasekar2018implicit,saxe_mathematical_2019,arora2019implicit} and understanding generalization dynamics~\cite{lampinen2018analytic, poggio2018theory,pmlr-v119-huh20a}.~\citet{Veness2021GatedLN, Budden2020GaussianGL} propose and study the Gated Linear Networks (GLNs) as a class of backpropagation-free neural architectures using geometric mixing. While GLNs appear similar to GDLNs, there are several differences. GDLN is a framework that schematizes how pathways of information flow impact learning dynamics within architecture and studies networks trained using back-propagation. Additionally, GLNs are good at online learning and continual learning, while in this work, we use the GDLN framework for understanding zero-shot generalization capabilities. In the GLN model, a neuron is defined as a gated geometric mixer of the output of linear networks in the previous layer, while in the GDLN model, the neurons are linear networks where the input is the output of the linear network along the previous path. In the GLN setup, multiple input-to-output connections (in successive layers) can be active for the same input, while in the GDLN setup, only one input-output connection (in successive layers) is active for one input.

In this work, we propose modeling the model architecture as a graph and study how pathways of information flow impact learning dynamics within an architecture. Previous works have also proposed analyzing neural networks as directed graphs using the Complex Network Theory~\cite{boccaletti2006complex}. \citet{scabini2021structure} analyzed the structure and performance of fully connected neural networks,~\citet{zambra2020emergence} focused on the \textit{emergence} of motifs in fully connected networks, \citet{testolin2020deep} study deep belief networks using techniques from Complex Network Theory literature and \citet{la2021characterizing} focused on convolution and fully connected networks, with ReLU non-linearity.

Our Gated Deep Linear Network framework is closely related to areas like~\textit{modular networks}~\citep{happel1994design, sharkey1997modularity, auda1999modular, johnson2017clevr, santoro2017simple},~\textit{routing networks}~\citep{rosenbaum2019routing} and~\textit{mixture of experts}~\citep{jacobs1991adaptive, jordan1994hierarchical, chen1999improved, yuksel2012twenty}. In these works, the common theme is to learn a set of \textit{modules} (or~\textit{experts}) that can be composed (or selected) using a \textit{controller} (or a router). The modules are generally instantiated as neural networks, while the controller can either be a neural network or a hand-designed policy. These approaches have been prominently used in natural language processing~\citep{shazeer2017outrageously, lepikhin2020gshard, fedus2021switch, lewis2021base}, computer vision~\citep{ahmed2016network, gross2017hard, yang2019condconv, wang2020deep} and reinforcement learning~\citep{multi-task-rl-with-soft-modularization, sodhani2021multi, andreas2017modular, goyal2019reinforcement, he2016opponent,Goyal2021RecurrentIM}.

Our work is also related to previous works in~\textbf{systematic generalization}~\cite{bahdanau2018systematic, bahdanau2019closure, lake2019compositional, ruis2020benchmark, gontier2020measuring} and multi-task learning~\cite{caruana1997multitask_learning, zhang2014facial_landmark_detection_by_deep_multitask_learning, kokkinos2017ubernet, radford2019language_models_are_unsupervised_multitask_learners, ruder2017overview, end_to_end_multi_task_learning_with_attention, mott2019towards, electronics9091363}. Specifically, we explore the role of model architecture and weight initialization on models' ability to exhibit systematic generalization and multi-task learning.

\section{Conclusion}
A key intuition in deep learning holds that a network's architecture influences learned representations, and should relate to task structure in order to achieve good performance and generalization. Here, we have introduced the Gated Deep Linear Network framework, which reveals how architecture--reflected by a simple nonlinear gating scheme along the edges of an architecture graph--controls pathways of information flow that govern learning dynamics, representation learning, and ultimately generalization. Our exact reductions and solutions show that learning dynamics take the form of a race, with greater representational reuse causing faster learning, imparting a bias toward shared representations. We validate our key insights on naturalistic datasets and with relaxed assumptions. An interesting future research direction will be to explore mechanisms for inferring the optimal architecture and gating for a given setup.

\section*{Acknowledgements}

We thank Hannah Sheahan, Timo Flesch, Devon Jarvis, and Olivier Delalleau for their feedback and comments. This work was supported by a Sir Henry Dale Fellowship from the Wellcome Trust and Royal Society (216386/Z/19/Z) to A.S., and the Sainsbury Wellcome Centre Core Grant from Wellcome (219627/Z/19/Z) and the Gatsby Charitable Foundation (GAT3755). A.S. is a CIFAR Azrieli Global Scholar in the Learning in Machines \& Brains program.

\bibliography{references}

\begin{thebibliography}{103}
\providecommand{\natexlab}[1]{#1}
\providecommand{\url}[1]{\texttt{#1}}
\expandafter\ifx\csname urlstyle\endcsname\relax
  \providecommand{\doi}[1]{doi: #1}\else
  \providecommand{\doi}{doi: \begingroup \urlstyle{rm}\Url}\fi

\bibitem[Ahmed et~al.(2016)Ahmed, Baig, and Torresani]{ahmed2016network}
Ahmed, K., Baig, M.~H., and Torresani, L.
\newblock Network of experts for large-scale image categorization.
\newblock In \emph{European Conference On Computer Vision}, pp.\  516--532.
  Springer, 2016.

\bibitem[Amodei et~al.(2016)Amodei, Ananthanarayanan, Anubhai, Bai, Battenberg,
  Case, Casper, Catanzaro, Chen, Chrzanowski, Coates, Diamos, Elsen, Engel,
  Fan, Fougner, Hannun, Jun, Han, LeGresley, Li, Lin, Narang, Ng, Ozair,
  Prenger, Qian, Raiman, Satheesh, Seetapun, Sengupta, Wang, Wang, Wang, Xiao,
  Xie, Yogatama, Zhan, and Zhu]{amodei2016deep}
Amodei, D., Ananthanarayanan, S., Anubhai, R., Bai, J., Battenberg, E., Case,
  C., Casper, J., Catanzaro, B., Chen, J., Chrzanowski, M., Coates, A., Diamos,
  G., Elsen, E., Engel, J.~H., Fan, L., Fougner, C., Hannun, A.~Y., Jun, B.,
  Han, T., LeGresley, P., Li, X., Lin, L., Narang, S., Ng, A.~Y., Ozair, S.,
  Prenger, R., Qian, S., Raiman, J., Satheesh, S., Seetapun, D., Sengupta, S.,
  Wang, C., Wang, Y., Wang, Z., Xiao, B., Xie, Y., Yogatama, D., Zhan, J., and
  Zhu, Z.
\newblock Deep speech 2 : End-to-end speech recognition in english and
  mandarin.
\newblock In Balcan, M. and Weinberger, K.~Q. (eds.), \emph{Proceedings of the
  33nd International Conference on Machine Learning, {ICML} 2016, New York
  City, NY, USA, June 19-24, 2016}, volume~48 of \emph{{JMLR} Workshop and
  Conference Proceedings}, pp.\  173--182. JMLR.org, 2016.
\newblock URL \url{http://proceedings.mlr.press/v48/amodei16.html}.

\bibitem[Andreas et~al.(2017)Andreas, Klein, and Levine]{andreas2017modular}
Andreas, J., Klein, D., and Levine, S.
\newblock Modular multitask reinforcement learning with policy sketches.
\newblock In Precup, D. and Teh, Y.~W. (eds.), \emph{Proceedings of the 34th
  International Conference on Machine Learning, {ICML} 2017, Sydney, NSW,
  Australia, 6-11 August 2017}, volume~70 of \emph{Proceedings of Machine
  Learning Research}, pp.\  166--175. {PMLR}, 2017.
\newblock URL \url{http://proceedings.mlr.press/v70/andreas17a.html}.

\bibitem[Arora et~al.(2020)Arora, Arora, Bruna, Cohen, Du, Ge, Gunasekar, Jin,
  Lee, Ma, and Others]{arora2020theory}
Arora, R., Arora, S., Bruna, J., Cohen, N., Du, S., Ge, R., Gunasekar, S., Jin,
  C., Lee, J., Ma, T., and Others.
\newblock Theory of deep learning, 2020.

\bibitem[Arora et~al.(2018)Arora, Cohen, and Hazan]{arora2018optimization}
Arora, S., Cohen, N., and Hazan, E.
\newblock On the optimization of deep networks: Implicit acceleration by
  overparameterization.
\newblock In Dy, J.~G. and Krause, A. (eds.), \emph{Proceedings of the 35th
  International Conference on Machine Learning, {ICML} 2018,
  Stockholmsm{\"{a}}ssan, Stockholm, Sweden, July 10-15, 2018}, volume~80 of
  \emph{Proceedings of Machine Learning Research}, pp.\  244--253. {PMLR},
  2018.
\newblock URL \url{http://proceedings.mlr.press/v80/arora18a.html}.

\bibitem[Arora et~al.(2019{\natexlab{a}})Arora, Cohen, Golowich, and
  Hu]{arora2018convergence}
Arora, S., Cohen, N., Golowich, N., and Hu, W.
\newblock A convergence analysis of gradient descent for deep linear neural
  networks.
\newblock In \emph{7th International Conference on Learning Representations,
  {ICLR} 2019, New Orleans, LA, USA, May 6-9, 2019}. OpenReview.net,
  2019{\natexlab{a}}.
\newblock URL \url{https://openreview.net/forum?id=SkMQg3C5K7}.

\bibitem[Arora et~al.(2019{\natexlab{b}})Arora, Cohen, Hu, and
  Luo]{arora2019implicit}
Arora, S., Cohen, N., Hu, W., and Luo, Y.
\newblock Implicit regularization in deep matrix factorization.
\newblock In Wallach, H.~M., Larochelle, H., Beygelzimer, A.,
  d'Alch{\'{e}}{-}Buc, F., Fox, E.~B., and Garnett, R. (eds.), \emph{Advances
  in Neural Information Processing Systems 32: Annual Conference on Neural
  Information Processing Systems 2019, NeurIPS 2019, December 8-14, 2019,
  Vancouver, BC, Canada}, pp.\  7411--7422, 2019{\natexlab{b}}.
\newblock URL
  \url{https://proceedings.neurips.cc/paper/2019/file/c0c783b5fc0d7d808f1d14a6e9c8280d-Paper.pdf}.

\bibitem[Arora et~al.(2019{\natexlab{c}})Arora, Du, Hu, Li, Salakhutdinov, and
  Wang]{arora_exact_2019}
Arora, S., Du, S.~S., Hu, W., Li, Z., Salakhutdinov, R., and Wang, R.
\newblock On exact computation with an infinitely wide neural net.
\newblock In Wallach, H.~M., Larochelle, H., Beygelzimer, A.,
  d'Alch{\'{e}}{-}Buc, F., Fox, E.~B., and Garnett, R. (eds.), \emph{Advances
  in Neural Information Processing Systems 32: Annual Conference on Neural
  Information Processing Systems 2019, NeurIPS 2019, December 8-14, 2019,
  Vancouver, BC, Canada}, pp.\  8139--8148, 2019{\natexlab{c}}.
\newblock URL
  \url{https://proceedings.neurips.cc/paper/2019/file/dbc4d84bfcfe2284ba11beffb853a8c4-Paper.pdf}.

\bibitem[Auda \& Kamel(1999)Auda and Kamel]{auda1999modular}
Auda, G. and Kamel, M.
\newblock Modular neural networks: A survey.
\newblock \emph{International Journal Of Neural Systems}, 9\penalty0
  (02):\penalty0 129--151, 1999.

\bibitem[Baevski et~al.(2020)Baevski, Zhou, Mohamed, and
  Auli]{baevski2020wav2vec}
Baevski, A., Zhou, Y., Mohamed, A., and Auli, M.
\newblock wav2vec 2.0: {A} framework for self-supervised learning of speech
  representations.
\newblock In Larochelle, H., Ranzato, M., Hadsell, R., Balcan, M., and Lin, H.
  (eds.), \emph{Advances in Neural Information Processing Systems 33: Annual
  Conference on Neural Information Processing Systems 2020, NeurIPS 2020,
  December 6-12, 2020, virtual}, 2020.
\newblock URL
  \url{https://proceedings.neurips.cc/paper/2020/file/92d1e1eb1cd6f9fba3227870bb6d7f07-Paper.pdf}.

\bibitem[Bahdanau et~al.(2019{\natexlab{a}})Bahdanau, de~Vries, O'Donnell,
  Murty, Beaudoin, Bengio, and Courville]{bahdanau2019closure}
Bahdanau, D., de~Vries, H., O'Donnell, T.~J., Murty, S., Beaudoin, P., Bengio,
  Y., and Courville, A.
\newblock Closure: Assessing systematic generalization of clevr models.
\newblock \emph{arXiv preprint arXiv:1912.05783}, 2019{\natexlab{a}}.

\bibitem[Bahdanau et~al.(2019{\natexlab{b}})Bahdanau, Murty, Noukhovitch,
  Nguyen, de~Vries, and Courville]{bahdanau2018systematic}
Bahdanau, D., Murty, S., Noukhovitch, M., Nguyen, T.~H., de~Vries, H., and
  Courville, A.~C.
\newblock Systematic generalization: What is required and can it be learned?
\newblock In \emph{7th International Conference on Learning Representations,
  {ICLR} 2019, New Orleans, LA, USA, May 6-9, 2019}. OpenReview.net,
  2019{\natexlab{b}}.
\newblock URL \url{https://openreview.net/forum?id=HkezXnA9YX}.

\bibitem[Bahri et~al.(2020)Bahri, Kadmon, Pennington, Schoenholz,
  Sohl-dickstein, and Ganguli]{bahri_statistical_2020}
Bahri, Y., Kadmon, J., Pennington, J., Schoenholz, S.~S., Sohl-dickstein, J.,
  and Ganguli, S.
\newblock Statistical {Mechanics} of {Deep} {Learning}.
\newblock \emph{Annual Review Of Condensed Matter Physics}, 11\penalty0
  (1):\penalty0 501--528, 2020.
\newblock \doi{10.1146/annurev-conmatphys-031119-050745}.
\newblock \_eprint: Https://doi.org/10.1146/annurev-conmatphys-031119-050745.

\bibitem[Baldi \& Hornik(1989)Baldi and Hornik]{Baldi1989}
Baldi, P. and Hornik, K.
\newblock Neural networks and principal component analysis: {Learning} from
  examples without local minima.
\newblock \emph{Neural Networks}, 2\penalty0 (1):\penalty0 53--58, 1989.
\newblock ISSN 08936080.
\newblock \doi{10.1016/0893-6080(89)90014-2}.
\newblock URL
  \url{http://linkinghub.elsevier.com/retrieve/pii/0893608089900142}.

\bibitem[Boccaletti et~al.(2006)Boccaletti, Latora, Moreno, Chavez, and
  Hwang]{boccaletti2006complex}
Boccaletti, S., Latora, V., Moreno, Y., Chavez, M., and Hwang, D.-u.
\newblock Complex networks: Structure and dynamics.
\newblock \emph{Physics Reports}, 424\penalty0 (4-5):\penalty0 175--308, 2006.

\bibitem[Budden et~al.(2020)Budden, Marblestone, Sezener, Lattimore, Wayne, and
  Veness]{Budden2020GaussianGL}
Budden, D., Marblestone, A.~H., Sezener, E., Lattimore, T., Wayne, G., and
  Veness, J.
\newblock Gaussian gated linear networks.
\newblock \emph{ArXiv}, abs/2006.05964, 2020.

\bibitem[Carleo et~al.(2019)Carleo, Cirac, Cranmer, Daudet, Schuld, Tishby,
  Vogt-maranto, and Zdeborov\'a]{Carleo2019}
Carleo, G., Cirac, I., Cranmer, K., Daudet, L., Schuld, M., Tishby, N.,
  Vogt-maranto, L., and Zdeborov\'a, L.
\newblock Machine learning and the physical sciences.
\newblock \emph{Rev. Mod. Phys.}, 91:\penalty0 045002, 2019.
\newblock \doi{10.1103/RevModPhys.91.045002}.

\bibitem[Caruana(1997)]{caruana1997multitask_learning}
Caruana, R.
\newblock Multitask learning.
\newblock \emph{Machine Learning}, 28\penalty0 (1):\penalty0 41--75, 1997.

\bibitem[Chen et~al.(1999)Chen, Xu, and Chi]{chen1999improved}
Chen, K., Xu, L., and Chi, H.
\newblock Improved learning algorithms for mixture of experts in multiclass
  classification.
\newblock \emph{Neural Networks}, 12\penalty0 (9):\penalty0 1229--1252, 1999.

\bibitem[Chizat et~al.(2019)Chizat, Oyallon, and Bach]{Chizat2018}
Chizat, L., Oyallon, E., and Bach, F.~R.
\newblock On lazy training in differentiable programming.
\newblock In Wallach, H.~M., Larochelle, H., Beygelzimer, A.,
  d'Alch{\'{e}}{-}Buc, F., Fox, E.~B., and Garnett, R. (eds.), \emph{Advances
  in Neural Information Processing Systems 32: Annual Conference on Neural
  Information Processing Systems 2019, NeurIPS 2019, December 8-14, 2019,
  Vancouver, BC, Canada}, pp.\  2933--2943, 2019.
\newblock URL
  \url{https://proceedings.neurips.cc/paper/2019/file/ae614c557843b1df326cb29c57225459-Paper.pdf}.

\bibitem[Collobert et~al.(2011)Collobert, Weston, Bottou, Karlen, Kavukcuoglu,
  and Kuksa]{collobert2011natural}
Collobert, R., Weston, J., Bottou, L., Karlen, M., Kavukcuoglu, K., and Kuksa,
  P.
\newblock Natural language processing (almost) from scratch.
\newblock \emph{Journal Of Machine Learning Research}, 12\penalty0
  (Article):\penalty0 2493--2537, 2011.

\bibitem[Combes et~al.(2018)Combes, Pezeshki, Shabanian, Courville, and
  Bengio]{des2018learning}
Combes, R.~T., Pezeshki, M., Shabanian, S., Courville, A.~C., and Bengio, Y.
\newblock On the learning dynamics of deep neural networks.
\newblock \emph{ArXiv}, abs/1809.06848, 2018.

\bibitem[Deng(2012)]{deng2012mnist}
Deng, L.
\newblock The mnist database of handwritten digit images for machine learning
  research.
\newblock \emph{Ieee Signal Processing Magazine}, 29\penalty0 (6):\penalty0
  141--142, 2012.

\bibitem[Du \& Hu(2019)Du and Hu]{du2019width}
Du, S.~S. and Hu, W.
\newblock Width provably matters in optimization for deep linear neural
  networks.
\newblock In Chaudhuri, K. and Salakhutdinov, R. (eds.), \emph{Proceedings of
  the 36th International Conference on Machine Learning, {ICML} 2019, 9-15 June
  2019, Long Beach, California, {USA}}, volume~97 of \emph{Proceedings of
  Machine Learning Research}, pp.\  1655--1664. {PMLR}, 2019.
\newblock URL \url{http://proceedings.mlr.press/v97/du19a.html}.

\bibitem[Fedus et~al.(2021)Fedus, Zoph, and Shazeer]{fedus2021switch}
Fedus, W., Zoph, B., and Shazeer, N.
\newblock Switch transformers: Scaling to trillion parameter models with simple
  and efficient sparsity.
\newblock \emph{ArXiv preprint}, abs/2101.03961, 2021.
\newblock URL \url{https://arxiv.org/abs/2101.03961}.

\bibitem[Flesch et~al.(2022)Flesch, Juechems, Dumbalska, Saxe, and
  Summerfield]{flesch_orthogonal_2022}
Flesch, T., Juechems, K., Dumbalska, T., Saxe, A., and Summerfield, C.
\newblock Orthogonal representations for robust context-dependent task
  performance in brains and neural networks.
\newblock \emph{Neuron}, 0\penalty0 (0), 2022.
\newblock ISSN 0896-6273.
\newblock \doi{10.1016/j.neuron.2022.01.005}.
\newblock URL \url{https://www.cell.com/neuron/abstract/S0896-6273(22)00005-8}.
\newblock Publisher: Elsevier.

\bibitem[Fort et~al.(2019)Fort, Hu, and Lakshminarayanan]{fort2019deep}
Fort, S., Hu, H., and Lakshminarayanan, B.
\newblock Deep ensembles: A loss landscape perspective.
\newblock \emph{ArXiv preprint}, abs/1912.02757, 2019.
\newblock URL \url{https://arxiv.org/abs/1912.02757}.

\bibitem[Fort et~al.(2020)Fort, Dziugaite, Paul, Kharaghani, Roy, and
  Ganguli]{fort2020deep}
Fort, S., Dziugaite, G.~K., Paul, M., Kharaghani, S., Roy, D.~M., and Ganguli,
  S.
\newblock Deep learning versus kernel learning: an empirical study of loss
  landscape geometry and the time evolution of the neural tangent kernel.
\newblock In Larochelle, H., Ranzato, M., Hadsell, R., Balcan, M., and Lin, H.
  (eds.), \emph{Advances in Neural Information Processing Systems 33: Annual
  Conference on Neural Information Processing Systems 2020, NeurIPS 2020,
  December 6-12, 2020, virtual}, 2020.
\newblock URL
  \url{https://proceedings.neurips.cc/paper/2020/file/405075699f065e43581f27d67bb68478-Paper.pdf}.

\bibitem[Fukumizu(1998)]{Fukumizu1998}
Fukumizu, K.
\newblock Effect of {Batch} {Learning} {In} {Multilayer} {Neural} {Networks}.
\newblock In \emph{Proceedings of the 5th {International} {Conference} on
  {Neural} {Information} {Processing}}, pp.\  67--70, 1998.

\bibitem[Girdhar et~al.(2022)Girdhar, Singh, Ravi, Van Der~Maaten, Joulin, and
  Misra]{girdhar2022omnivore}
Girdhar, R., Singh, M., Ravi, N., Van Der~Maaten, L., Joulin, A., and Misra, I.
\newblock Omnivore: A single model for many visual modalities.
\newblock \emph{ArXiv preprint}, abs/2201.08377, 2022.
\newblock URL \url{https://arxiv.org/abs/2201.08377}.

\bibitem[Gontier et~al.(2020)Gontier, Sinha, Reddy, and
  Pal]{gontier2020measuring}
Gontier, N., Sinha, K., Reddy, S., and Pal, C.
\newblock Measuring systematic generalization in neural proof generation with
  transformers.
\newblock In Larochelle, H., Ranzato, M., Hadsell, R., Balcan, M., and Lin, H.
  (eds.), \emph{Advances in Neural Information Processing Systems 33: Annual
  Conference on Neural Information Processing Systems 2020, NeurIPS 2020,
  December 6-12, 2020, virtual}, 2020.
\newblock URL
  \url{https://proceedings.neurips.cc/paper/2020/file/fc84ad56f9f547eb89c72b9bac209312-Paper.pdf}.

\bibitem[Goyal et~al.(2020)Goyal, Sodhani, Binas, Peng, Levine, and
  Bengio]{goyal2019reinforcement}
Goyal, A., Sodhani, S., Binas, J., Peng, X.~B., Levine, S., and Bengio, Y.
\newblock Reinforcement learning with competitive ensembles of
  information-constrained primitives.
\newblock In \emph{8th International Conference on Learning Representations,
  {ICLR} 2020, Addis Ababa, Ethiopia, April 26-30, 2020}. OpenReview.net, 2020.
\newblock URL \url{https://openreview.net/forum?id=ryxgJTEYDr}.

\bibitem[Goyal et~al.(2021)Goyal, Lamb, Hoffmann, Sodhani, Levine, Bengio, and
  Sch{\"o}lkopf]{Goyal2021RecurrentIM}
Goyal, A., Lamb, A., Hoffmann, J., Sodhani, S., Levine, S., Bengio, Y., and
  Sch{\"o}lkopf, B.
\newblock Recurrent independent mechanisms.
\newblock \emph{ArXiv}, abs/1909.10893, 2021.

\bibitem[Gross et~al.(2017)Gross, Ranzato, and Szlam]{gross2017hard}
Gross, S., Ranzato, M., and Szlam, A.
\newblock Hard mixtures of experts for large scale weakly supervised vision.
\newblock In \emph{2017 {IEEE} Conference on Computer Vision and Pattern
  Recognition, {CVPR} 2017, Honolulu, HI, USA, July 21-26, 2017}, pp.\
  5085--5093. {IEEE} Computer Society, 2017.
\newblock \doi{10.1109/CVPR.2017.540}.
\newblock URL \url{https://doi.org/10.1109/CVPR.2017.540}.

\bibitem[Gunasekar et~al.(2018)Gunasekar, Lee, Soudry, and
  Srebro]{gunasekar2018implicit}
Gunasekar, S., Lee, J.~D., Soudry, D., and Srebro, N.
\newblock Implicit bias of gradient descent on linear convolutional networks.
\newblock In Bengio, S., Wallach, H.~M., Larochelle, H., Grauman, K.,
  Cesa{-}Bianchi, N., and Garnett, R. (eds.), \emph{Advances in Neural
  Information Processing Systems 31: Annual Conference on Neural Information
  Processing Systems 2018, NeurIPS 2018, December 3-8, 2018, Montr{\'{e}}al,
  Canada}, pp.\  9482--9491, 2018.
\newblock URL
  \url{https://proceedings.neurips.cc/paper/2018/file/0e98aeeb54acf612b9eb4e48a269814c-Paper.pdf}.

\bibitem[Happel \& Murre(1994)Happel and Murre]{happel1994design}
Happel, B.~L. and Murre, J.~M.
\newblock Design and evolution of modular neural network architectures.
\newblock \emph{Neural Networks}, 7\penalty0 (6-7):\penalty0 985--1004, 1994.

\bibitem[Harris et~al.(2020)Harris, Millman, van~der Walt, Gommers, Virtanen,
  Cournapeau, Wieser, Taylor, Berg, Smith, Kern, Picus, Hoyer, van Kerkwijk,
  Brett, Haldane, del R{'{\i}}o, Wiebe, Peterson, G{'{e}}rard-Marchant,
  Sheppard, Reddy, Weckesser, Abbasi, Gohlke, and Oliphant]{harris2020array}
Harris, C.~R., Millman, K.~J., van~der Walt, S.~J., Gommers, R., Virtanen, P.,
  Cournapeau, D., Wieser, E., Taylor, J., Berg, S., Smith, N.~J., Kern, R.,
  Picus, M., Hoyer, S., van Kerkwijk, M.~H., Brett, M., Haldane, A., del
  R{'{\i}}o, J.~F., Wiebe, M., Peterson, P., G{'{e}}rard-Marchant, P.,
  Sheppard, K., Reddy, T., Weckesser, W., Abbasi, H., Gohlke, C., and Oliphant,
  T.~E.
\newblock Array programming with {NumPy}.
\newblock \emph{Nature}, 585\penalty0 (7825):\penalty0 357--362, 2020.
\newblock \doi{10.1038/s41586-020-2649-2}.
\newblock URL \url{https://doi.org/10.1038/s41586-020-2649-2}.

\bibitem[He \& Boyd{-}Graber(2016)He and Boyd{-}Graber]{he2016opponent}
He, H. and Boyd{-}Graber, J.~L.
\newblock Opponent modeling in deep reinforcement learning.
\newblock In Balcan, M. and Weinberger, K.~Q. (eds.), \emph{Proceedings of the
  33nd International Conference on Machine Learning, {ICML} 2016, New York
  City, NY, USA, June 19-24, 2016}, volume~48 of \emph{{JMLR} Workshop and
  Conference Proceedings}, pp.\  1804--1813. JMLR.org, 2016.
\newblock URL \url{http://proceedings.mlr.press/v48/he16.html}.

\bibitem[He et~al.(2016)He, Zhang, Ren, and Sun]{resnet}
He, K., Zhang, X., Ren, S., and Sun, J.
\newblock Deep residual learning for image recognition.
\newblock In \emph{2016 {IEEE} Conference on Computer Vision and Pattern
  Recognition, {CVPR} 2016, Las Vegas, NV, USA, June 27-30, 2016}, pp.\
  770--778. {IEEE} Computer Society, 2016.
\newblock \doi{10.1109/CVPR.2016.90}.
\newblock URL \url{https://doi.org/10.1109/CVPR.2016.90}.

\bibitem[Huh(2020)]{pmlr-v119-huh20a}
Huh, D.
\newblock Curvature-corrected learning dynamics in deep neural networks.
\newblock In \emph{Proceedings of the 37th International Conference on Machine
  Learning, {ICML} 2020, 13-18 July 2020, Virtual Event}, volume 119 of
  \emph{Proceedings of Machine Learning Research}, pp.\  4552--4560. {PMLR},
  2020.
\newblock URL \url{http://proceedings.mlr.press/v119/huh20a.html}.

\bibitem[Idelbayev(2020)]{Idelbayev18a}
Idelbayev, Y.
\newblock Proper {ResNet} implementation for {CIFAR10/CIFAR100} in {PyTorch}.
\newblock \url{https://github.com/akamaster/pytorch\_resnet\_cifar10}, 2020.

\bibitem[Jacobs et~al.(1991)Jacobs, Jordan, Nowlan, and
  Hinton]{jacobs1991adaptive}
Jacobs, R.~A., Jordan, M.~I., Nowlan, S.~J., and Hinton, G.~E.
\newblock Adaptive mixtures of local experts.
\newblock \emph{Neural Computation}, 3\penalty0 (1):\penalty0 79--87, 1991.

\bibitem[Jacot et~al.(2018)Jacot, Hongler, and Gabriel]{jacot2018neural}
Jacot, A., Hongler, C., and Gabriel, F.
\newblock Neural tangent kernel: Convergence and generalization in neural
  networks.
\newblock In Bengio, S., Wallach, H.~M., Larochelle, H., Grauman, K.,
  Cesa{-}Bianchi, N., and Garnett, R. (eds.), \emph{Advances in Neural
  Information Processing Systems 31: Annual Conference on Neural Information
  Processing Systems 2018, NeurIPS 2018, December 3-8, 2018, Montr{\'{e}}al,
  Canada}, pp.\  8580--8589, 2018.
\newblock URL
  \url{https://proceedings.neurips.cc/paper/2018/file/5a4be1fa34e62bb8a6ec6b91d2462f5a-Paper.pdf}.

\bibitem[Ji \& Telgarsky(2019)Ji and Telgarsky]{ji2018gradient}
Ji, Z. and Telgarsky, M.
\newblock Gradient descent aligns the layers of deep linear networks.
\newblock In \emph{7th International Conference on Learning Representations,
  {ICLR} 2019, New Orleans, LA, USA, May 6-9, 2019}. OpenReview.net, 2019.
\newblock URL \url{https://openreview.net/forum?id=HJflg30qKX}.

\bibitem[Johnson et~al.(2017)Johnson, Hariharan, van~der Maaten, Fei{-}Fei,
  Zitnick, and Girshick]{johnson2017clevr}
Johnson, J., Hariharan, B., van~der Maaten, L., Fei{-}Fei, L., Zitnick, C.~L.,
  and Girshick, R.~B.
\newblock {CLEVR:} {A} diagnostic dataset for compositional language and
  elementary visual reasoning.
\newblock In \emph{2017 {IEEE} Conference on Computer Vision and Pattern
  Recognition, {CVPR} 2017, Honolulu, HI, USA, July 21-26, 2017}, pp.\
  1988--1997. {IEEE} Computer Society, 2017.
\newblock \doi{10.1109/CVPR.2017.215}.
\newblock URL \url{https://doi.org/10.1109/CVPR.2017.215}.

\bibitem[Jordan \& Jacobs(1994)Jordan and Jacobs]{jordan1994hierarchical}
Jordan, M.~I. and Jacobs, R.~A.
\newblock Hierarchical mixtures of experts and the em algorithm.
\newblock \emph{Neural Computation}, 6\penalty0 (2):\penalty0 181--214, 1994.

\bibitem[Kokkinos(2017)]{kokkinos2017ubernet}
Kokkinos, I.
\newblock Ubernet: Training a universal convolutional neural network for low-,
  mid-, and high-level vision using diverse datasets and limited memory.
\newblock In \emph{2017 {IEEE} Conference on Computer Vision and Pattern
  Recognition, {CVPR} 2017, Honolulu, HI, USA, July 21-26, 2017}, pp.\
  5454--5463. {IEEE} Computer Society, 2017.
\newblock \doi{10.1109/CVPR.2017.579}.
\newblock URL \url{https://doi.org/10.1109/CVPR.2017.579}.

\bibitem[Krizhevsky et~al.(2009)Krizhevsky, Hinton, and
  Others]{krizhevsky2009learning}
Krizhevsky, A., Hinton, G., and Others.
\newblock Learning multiple layers of features from tiny images.
\newblock 2009.

\bibitem[La~Malfa et~al.(2021)La~Malfa, La~Malfa, Nicosia, and
  Latora]{la2021characterizing}
La~Malfa, E., La~Malfa, G., Nicosia, G., and Latora, V.
\newblock Characterizing learning dynamics of deep neural networks via complex
  networks.
\newblock In \emph{2021 Ieee 33rd International Conference On Tools With
  Artificial Intelligence (ictai)}, pp.\  344--351. Ieee, 2021.

\bibitem[Lake(2019)]{lake2019compositional}
Lake, B.~M.
\newblock Compositional generalization through meta sequence-to-sequence
  learning.
\newblock In Wallach, H.~M., Larochelle, H., Beygelzimer, A.,
  d'Alch{\'{e}}{-}Buc, F., Fox, E.~B., and Garnett, R. (eds.), \emph{Advances
  in Neural Information Processing Systems 32: Annual Conference on Neural
  Information Processing Systems 2019, NeurIPS 2019, December 8-14, 2019,
  Vancouver, BC, Canada}, pp.\  9788--9798, 2019.
\newblock URL
  \url{https://proceedings.neurips.cc/paper/2019/file/f4d0e2e7fc057a58f7ca4a391f01940a-Paper.pdf}.

\bibitem[Lampinen \& Ganguli(2019)Lampinen and Ganguli]{lampinen2018analytic}
Lampinen, A.~K. and Ganguli, S.
\newblock An analytic theory of generalization dynamics and transfer learning
  in deep linear networks.
\newblock In \emph{7th International Conference on Learning Representations,
  {ICLR} 2019, New Orleans, LA, USA, May 6-9, 2019}. OpenReview.net, 2019.
\newblock URL \url{https://openreview.net/forum?id=ryfMLoCqtQ}.

\bibitem[Laurent \& von Brecht(2018)Laurent and von Brecht]{laurent2018deep}
Laurent, T. and von Brecht, J.
\newblock Deep linear networks with arbitrary loss: All local minima are
  global.
\newblock In Dy, J.~G. and Krause, A. (eds.), \emph{Proceedings of the 35th
  International Conference on Machine Learning, {ICML} 2018,
  Stockholmsm{\"{a}}ssan, Stockholm, Sweden, July 10-15, 2018}, volume~80 of
  \emph{Proceedings of Machine Learning Research}, pp.\  2908--2913. {PMLR},
  2018.
\newblock URL \url{http://proceedings.mlr.press/v80/laurent18a.html}.

\bibitem[Lee et~al.(2019)Lee, Xiao, Schoenholz, Bahri, Novak, Sohl{-}Dickstein,
  and Pennington]{Lee2019}
Lee, J., Xiao, L., Schoenholz, S.~S., Bahri, Y., Novak, R., Sohl{-}Dickstein,
  J., and Pennington, J.
\newblock Wide neural networks of any depth evolve as linear models under
  gradient descent.
\newblock In Wallach, H.~M., Larochelle, H., Beygelzimer, A.,
  d'Alch{\'{e}}{-}Buc, F., Fox, E.~B., and Garnett, R. (eds.), \emph{Advances
  in Neural Information Processing Systems 32: Annual Conference on Neural
  Information Processing Systems 2019, NeurIPS 2019, December 8-14, 2019,
  Vancouver, BC, Canada}, pp.\  8570--8581, 2019.
\newblock URL
  \url{https://proceedings.neurips.cc/paper/2019/file/0d1a9651497a38d8b1c3871c84528bd4-Paper.pdf}.

\bibitem[Lepikhin et~al.(2020)Lepikhin, Lee, Xu, Chen, Firat, Huang, Krikun,
  Shazeer, and Chen]{lepikhin2020gshard}
Lepikhin, D., Lee, H., Xu, Y., Chen, D., Firat, O., Huang, Y., Krikun, M.,
  Shazeer, N., and Chen, Z.
\newblock Gshard: Scaling giant models with conditional computation and
  automatic sharding.
\newblock In \emph{International Conference on Learning Representations}, 2020.

\bibitem[Lewis et~al.(2021)Lewis, Bhosale, Dettmers, Goyal, and
  Zettlemoyer]{lewis2021base}
Lewis, M., Bhosale, S., Dettmers, T., Goyal, N., and Zettlemoyer, L.
\newblock Base layers: Simplifying training of large, sparse models.
\newblock \emph{ArXiv preprint}, abs/2103.16716, 2021.
\newblock URL \url{https://arxiv.org/abs/2103.16716}.

\bibitem[Liu et~al.(2019{\natexlab{a}})Liu, Johns, and
  Davison]{end_to_end_multi_task_learning_with_attention}
Liu, S., Johns, E., and Davison, A.~J.
\newblock End-to-end multi-task learning with attention.
\newblock In \emph{{IEEE} Conference on Computer Vision and Pattern
  Recognition, {CVPR} 2019, Long Beach, CA, USA, June 16-20, 2019}, pp.\
  1871--1880. Computer Vision Foundation / {IEEE}, 2019{\natexlab{a}}.
\newblock \doi{10.1109/CVPR.2019.00197}.
\newblock URL
  \url{http://openaccess.thecvf.com/content\_CVPR\_2019/html/Liu\_End-To-End\_Multi-Task\_Learning\_With\_Attention\_CVPR\_2019\_paper.html}.

\bibitem[Liu et~al.(2019{\natexlab{b}})Liu, He, Chen, and Gao]{liu2019multi}
Liu, X., He, P., Chen, W., and Gao, J.
\newblock Multi-task deep neural networks for natural language understanding.
\newblock In \emph{Proceedings of the 57th Annual Meeting of the Association
  for Computational Linguistics}, pp.\  4487--4496, Florence, Italy,
  2019{\natexlab{b}}. Association for Computational Linguistics.
\newblock \doi{10.18653/v1/P19-1441}.
\newblock URL \url{https://aclanthology.org/P19-1441}.

\bibitem[Mei et~al.(2018)Mei, Montanari, and Nguyen]{mei_mean_2018}
Mei, S., Montanari, A., and Nguyen, P.-m.
\newblock A mean field view of the landscape of two-layer neural networks.
\newblock \emph{Proceedings Of The National Academy Of Sciences}, 115\penalty0
  (33):\penalty0 E7665--e7671, 2018.
\newblock ISSN 0027-8424, 1091-6490.
\newblock \doi{10.1073/pnas.1806579115}.
\newblock URL \url{https://www.pnas.org/content/115/33/E7665}.
\newblock Publisher: National Academy Of Sciences Section: Pnas Plus.

\bibitem[Mnih et~al.(2015)Mnih, Kavukcuoglu, Silver, Rusu, Veness, Bellemare,
  Graves, Riedmiller, Fidjeland, Ostrovski, and Others]{nature_dqn}
Mnih, V., Kavukcuoglu, K., Silver, D., Rusu, A.~A., Veness, J., Bellemare,
  M.~G., Graves, A., Riedmiller, M., Fidjeland, A.~K., Ostrovski, G., and
  Others.
\newblock Human-level control through deep reinforcement learning.
\newblock \emph{Nature}, 518\penalty0 (7540):\penalty0 529--533, 2015.

\bibitem[Mott et~al.(2019)Mott, Zoran, Chrzanowski, Wierstra, and
  Rezende]{mott2019towards}
Mott, A., Zoran, D., Chrzanowski, M., Wierstra, D., and Rezende, D.~J.
\newblock Towards interpretable reinforcement learning using attention
  augmented agents.
\newblock In Wallach, H.~M., Larochelle, H., Beygelzimer, A.,
  d'Alch{\'{e}}{-}Buc, F., Fox, E.~B., and Garnett, R. (eds.), \emph{Advances
  in Neural Information Processing Systems 32: Annual Conference on Neural
  Information Processing Systems 2019, NeurIPS 2019, December 8-14, 2019,
  Vancouver, BC, Canada}, pp.\  12329--12338, 2019.
\newblock URL
  \url{https://proceedings.neurips.cc/paper/2019/file/e9510081ac30ffa83f10b68cde1cac07-Paper.pdf}.

\bibitem[Ngiam et~al.(2011)Ngiam, Khosla, Kim, Nam, Lee, and
  Ng]{ngiam2011multimodal}
Ngiam, J., Khosla, A., Kim, M., Nam, J., Lee, H., and Ng, A.~Y.
\newblock Multimodal deep learning.
\newblock In Getoor, L. and Scheffer, T. (eds.), \emph{Proceedings of the 28th
  International Conference on Machine Learning, {ICML} 2011, Bellevue,
  Washington, USA, June 28 - July 2, 2011}, pp.\  689--696. Omnipress, 2011.
\newblock URL \url{https://icml.cc/2011/papers/399\_icmlpaper.pdf}.

\bibitem[Nguyen et~al.(2020)Nguyen, Raghu, and Kornblith]{nguyen2020wide}
Nguyen, T., Raghu, M., and Kornblith, S.
\newblock Do wide and deep networks learn the same things? uncovering how
  neural network representations vary with width and depth.
\newblock In \emph{International Conference on Learning Representations}, 2020.

\bibitem[Paszke et~al.(2019)Paszke, Gross, Massa, Lerer, Bradbury, Chanan,
  Killeen, Lin, Gimelshein, Antiga, Desmaison, K{\"{o}}pf, Yang, DeVito,
  Raison, Tejani, Chilamkurthy, Steiner, Fang, Bai, and Chintala]{2019pytorch}
Paszke, A., Gross, S., Massa, F., Lerer, A., Bradbury, J., Chanan, G., Killeen,
  T., Lin, Z., Gimelshein, N., Antiga, L., Desmaison, A., K{\"{o}}pf, A., Yang,
  E., DeVito, Z., Raison, M., Tejani, A., Chilamkurthy, S., Steiner, B., Fang,
  L., Bai, J., and Chintala, S.
\newblock Pytorch: An imperative style, high-performance deep learning library.
\newblock In Wallach, H.~M., Larochelle, H., Beygelzimer, A.,
  d'Alch{\'{e}}{-}Buc, F., Fox, E.~B., and Garnett, R. (eds.), \emph{Advances
  in Neural Information Processing Systems 32: Annual Conference on Neural
  Information Processing Systems 2019, NeurIPS 2019, December 8-14, 2019,
  Vancouver, BC, Canada}, pp.\  8024--8035, 2019.
\newblock URL
  \url{https://proceedings.neurips.cc/paper/2019/file/bdbca288fee7f92f2bfa9f7012727740-Paper.pdf}.

\bibitem[Patel et~al.(2015)Patel, Nguyen, and Baraniuk]{patel2015probabilistic}
Patel, A.~B., Nguyen, T., and Baraniuk, R.~G.
\newblock A probabilistic theory of deep learning.
\newblock \emph{ArXiv preprint}, abs/1504.00641, 2015.
\newblock URL \url{https://arxiv.org/abs/1504.00641}.

\bibitem[Poggio et~al.(2018)Poggio, Liao, Miranda, Burbanski, and
  Hidary]{poggio2018theory}
Poggio, T., Liao, Q., Miranda, B., Burbanski, A., and Hidary, J.
\newblock Theory iiib: Generalization in deep networks.
\newblock Technical report, Center for Brains, Minds and Machines (CBMM),
  arXiv. org, 2018.

\bibitem[Radford et~al.(2019)Radford, Wu, Child, Luan, Amodei, and
  Sutskever]{radford2019language_models_are_unsupervised_multitask_learners}
Radford, A., Wu, J., Child, R., Luan, D., Amodei, D., and Sutskever, I.
\newblock Language models are unsupervised multitask learners.
\newblock \emph{Openai Blog}, 1\penalty0 (8):\penalty0 9, 2019.

\bibitem[Raghu et~al.(2017)Raghu, Poole, Kleinberg, Ganguli, and
  Sohl{-}Dickstein]{raghu2017expressive}
Raghu, M., Poole, B., Kleinberg, J.~M., Ganguli, S., and Sohl{-}Dickstein, J.
\newblock On the expressive power of deep neural networks.
\newblock In Precup, D. and Teh, Y.~W. (eds.), \emph{Proceedings of the 34th
  International Conference on Machine Learning, {ICML} 2017, Sydney, NSW,
  Australia, 6-11 August 2017}, volume~70 of \emph{Proceedings of Machine
  Learning Research}, pp.\  2847--2854. {PMLR}, 2017.
\newblock URL \url{http://proceedings.mlr.press/v70/raghu17a.html}.

\bibitem[Roberts et~al.(2021)Roberts, Yaida, and Hanin]{roberts2021principles}
Roberts, D.~A., Yaida, S., and Hanin, B.
\newblock The principles of deep learning theory.
\newblock \emph{ArXiv preprint}, abs/2106.10165, 2021.
\newblock URL \url{https://arxiv.org/abs/2106.10165}.

\bibitem[Rosenbaum et~al.(2019)Rosenbaum, Cases, Riemer, and
  Klinger]{rosenbaum2019routing}
Rosenbaum, C., Cases, I., Riemer, M., and Klinger, T.
\newblock Routing networks and the challenges of modular and compositional
  computation.
\newblock \emph{ArXiv preprint}, abs/1904.12774, 2019.
\newblock URL \url{https://arxiv.org/abs/1904.12774}.

\bibitem[Rotskoff \& Vanden{-}Eijnden(2018)Rotskoff and
  Vanden{-}Eijnden]{rotskoff_parameters_2018}
Rotskoff, G.~M. and Vanden{-}Eijnden, E.
\newblock Parameters as interacting particles: long time convergence and
  asymptotic error scaling of neural networks.
\newblock In Bengio, S., Wallach, H.~M., Larochelle, H., Grauman, K.,
  Cesa{-}Bianchi, N., and Garnett, R. (eds.), \emph{Advances in Neural
  Information Processing Systems 31: Annual Conference on Neural Information
  Processing Systems 2018, NeurIPS 2018, December 3-8, 2018, Montr{\'{e}}al,
  Canada}, pp.\  7146--7155, 2018.
\newblock URL
  \url{https://proceedings.neurips.cc/paper/2018/file/196f5641aa9dc87067da4ff90fd81e7b-Paper.pdf}.

\bibitem[Ruder(2017)]{ruder2017overview}
Ruder, S.
\newblock An overview of multi-task learning in deep neural networks.
\newblock \emph{ArXiv preprint}, abs/1706.05098, 2017.
\newblock URL \url{https://arxiv.org/abs/1706.05098}.

\bibitem[Ruis et~al.(2020)Ruis, Andreas, Baroni, Bouchacourt, and
  Lake]{ruis2020benchmark}
Ruis, L., Andreas, J., Baroni, M., Bouchacourt, D., and Lake, B.~M.
\newblock A benchmark for systematic generalization in grounded language
  understanding.
\newblock \emph{Advances in Neural Information Processing Systems}, 33, 2020.

\bibitem[Salakhutdinov(2014{\natexlab{a}})]{goodfellow2016deep}
Salakhutdinov, R.
\newblock Deep learning.
\newblock In Macskassy, S.~A., Perlich, C., Leskovec, J., Wang, W., and Ghani,
  R. (eds.), \emph{The 20th {ACM} {SIGKDD} International Conference on
  Knowledge Discovery and Data Mining, {KDD} '14, New York, NY, {USA} - August
  24 - 27, 2014}, pp.\  1973. {ACM}, 2014{\natexlab{a}}.
\newblock \doi{10.1145/2623330.2630809}.
\newblock URL \url{https://doi.org/10.1145/2623330.2630809}.

\bibitem[Salakhutdinov(2014{\natexlab{b}})]{lecun2015deep}
Salakhutdinov, R.
\newblock Deep learning.
\newblock In Macskassy, S.~A., Perlich, C., Leskovec, J., Wang, W., and Ghani,
  R. (eds.), \emph{The 20th {ACM} {SIGKDD} International Conference on
  Knowledge Discovery and Data Mining, {KDD} '14, New York, NY, {USA} - August
  24 - 27, 2014}, pp.\  1973. {ACM}, 2014{\natexlab{b}}.
\newblock \doi{10.1145/2623330.2630809}.
\newblock URL \url{https://doi.org/10.1145/2623330.2630809}.

\bibitem[Santoro et~al.(2017)Santoro, Raposo, Barrett, Malinowski, Pascanu,
  Battaglia, and Lillicrap]{santoro2017simple}
Santoro, A., Raposo, D., Barrett, D. G.~T., Malinowski, M., Pascanu, R.,
  Battaglia, P.~W., and Lillicrap, T.
\newblock A simple neural network module for relational reasoning.
\newblock In Guyon, I., von Luxburg, U., Bengio, S., Wallach, H.~M., Fergus,
  R., Vishwanathan, S. V.~N., and Garnett, R. (eds.), \emph{Advances in Neural
  Information Processing Systems 30: Annual Conference on Neural Information
  Processing Systems 2017, December 4-9, 2017, Long Beach, CA, {USA}}, pp.\
  4967--4976, 2017.
\newblock URL
  \url{https://proceedings.neurips.cc/paper/2017/file/e6acf4b0f69f6f6e60e9a815938aa1ff-Paper.pdf}.

\bibitem[Saxe et~al.(2014)Saxe, McClelland, and Ganguli]{Saxe2014}
Saxe, A.~M., McClelland, J.~L., and Ganguli, S.
\newblock Exact solutions to the nonlinear dynamics of learning in deep linear
  neural networks.
\newblock In Bengio, Y. and LeCun, Y. (eds.), \emph{2nd International
  Conference on Learning Representations, {ICLR} 2014, Banff, AB, Canada, April
  14-16, 2014, Conference Track Proceedings}, 2014.
\newblock URL \url{http://arxiv.org/abs/1312.6120}.

\bibitem[Saxe et~al.(2019)Saxe, Mcclelland, and
  Ganguli]{saxe_mathematical_2019}
Saxe, A.~M., Mcclelland, J.~L., and Ganguli, S.
\newblock A mathematical theory of semantic development in deep neural
  networks.
\newblock \emph{Proceedings Of The National Academy Of Sciences}, 116\penalty0
  (23):\penalty0 11537--11546, 2019.
\newblock ISSN 0027-8424, 1091-6490.
\newblock \doi{10.1073/pnas.1820226116}.

\bibitem[Scabini \& Bruno(2021)Scabini and Bruno]{scabini2021structure}
Scabini, L.~F. and Bruno, O.~M.
\newblock Structure and performance of fully connected neural networks:
  Emerging complex network properties.
\newblock \emph{ArXiv preprint}, abs/2107.14062, 2021.
\newblock URL \url{https://arxiv.org/abs/2107.14062}.

\bibitem[Sharkey(1997)]{sharkey1997modularity}
Sharkey, A. J.~C.
\newblock Modularity, combining and artificial neural nets.
\newblock \emph{Connection Science}, 9\penalty0 (1):\penalty0 3--10, 1997.

\bibitem[Shazeer et~al.(2017)Shazeer, Mirhoseini, Maziarz, Davis, Le, Hinton,
  and Dean]{shazeer2017outrageously}
Shazeer, N., Mirhoseini, A., Maziarz, K., Davis, A., Le, Q.~V., Hinton, G.~E.,
  and Dean, J.
\newblock Outrageously large neural networks: The sparsely-gated
  mixture-of-experts layer.
\newblock In \emph{5th International Conference on Learning Representations,
  {ICLR} 2017, Toulon, France, April 24-26, 2017, Conference Track
  Proceedings}. OpenReview.net, 2017.
\newblock URL \url{https://openreview.net/forum?id=B1ckMDqlg}.

\bibitem[Silver et~al.(2017)Silver, Schrittwieser, Simonyan, Antonoglou, Huang,
  Guez, Hubert, Baker, Lai, Bolton, and Others]{alpha_zero}
Silver, D., Schrittwieser, J., Simonyan, K., Antonoglou, I., Huang, A., Guez,
  A., Hubert, T., Baker, L., Lai, M., Bolton, A., and Others.
\newblock Mastering the game of go without human knowledge.
\newblock \emph{Nature}, 550\penalty0 (7676):\penalty0 354--359, 2017.

\bibitem[Sinha et~al.(2019)Sinha, Sodhani, Dong, Pineau, and
  Hamilton]{Sinha2019CLUTRRAD}
Sinha, K., Sodhani, S., Dong, J., Pineau, J., and Hamilton, W.~L.
\newblock {CLUTRR}: A diagnostic benchmark for inductive reasoning from text.
\newblock In \emph{Proceedings of the 2019 Conference on Empirical Methods in
  Natural Language Processing and the 9th International Joint Conference on
  Natural Language Processing (EMNLP-IJCNLP)}, pp.\  4506--4515, Hong Kong,
  China, 2019. Association for Computational Linguistics.
\newblock \doi{10.18653/v1/D19-1458}.
\newblock URL \url{https://aclanthology.org/D19-1458}.

\bibitem[Sinha et~al.(2020)Sinha, Sodhani, Pineau, and
  Hamilton]{Sinha2020EvaluatingLG}
Sinha, K., Sodhani, S., Pineau, J., and Hamilton, W.~L.
\newblock Evaluating logical generalization in graph neural networks.
\newblock \emph{ArXiv preprint}, abs/2003.06560, 2020.
\newblock URL \url{https://arxiv.org/abs/2003.06560}.

\bibitem[Sirignano \& Spiliopoulos(2020)Sirignano and
  Spiliopoulos]{sirignano_mean_2020}
Sirignano, J. and Spiliopoulos, K.
\newblock Mean field analysis of neural networks: {A} central limit theorem.
\newblock \emph{Stochastic Processes And Their Applications}, 130\penalty0
  (3):\penalty0 1820--1852, 2020.
\newblock ISSN 0304-4149.
\newblock \doi{10.1016/j.spa.2019.06.003}.
\newblock URL
  \url{https://www.sciencedirect.com/science/article/pii/S0304414918306197}.

\bibitem[Sodhani(2022)]{Sodhani_xplogger_Logging_utility_2022}
Sodhani, S.
\newblock {xplogger: Logging utility for ML experiments}, 2 2022.
\newblock URL \url{https://github.com/shagunsodhani/xplogger}.

\bibitem[Sodhani et~al.(2021)Sodhani, Zhang, and Pineau]{sodhani2021multi}
Sodhani, S., Zhang, A., and Pineau, J.
\newblock Multi-task reinforcement learning with context-based representations.
\newblock In \emph{International Conference On Machine Learning (icml)}, 2021.

\bibitem[Standley et~al.(2020)Standley, Zamir, Chen, Guibas, Malik, and
  Savarese]{standley2020tasks}
Standley, T., Zamir, A.~R., Chen, D., Guibas, L.~J., Malik, J., and Savarese,
  S.
\newblock Which tasks should be learned together in multi-task learning?
\newblock In \emph{Proceedings of the 37th International Conference on Machine
  Learning, {ICML} 2020, 13-18 July 2020, Virtual Event}, volume 119 of
  \emph{Proceedings of Machine Learning Research}, pp.\  9120--9132. {PMLR},
  2020.
\newblock URL \url{http://proceedings.mlr.press/v119/standley20a.html}.

\bibitem[Straat \& Biehl(2019)Straat and Biehl]{straat2019line}
Straat, M. and Biehl, M.
\newblock On-line learning dynamics of relu neural networks using statistical
  physics techniques.
\newblock \emph{ArXiv preprint}, abs/1903.07378, 2019.
\newblock URL \url{https://arxiv.org/abs/1903.07378}.

\bibitem[Team(2020)]{reback2020pandas}
Team, T. P.~D.
\newblock pandas-dev/pandas: Pandas, 2020.
\newblock URL \url{https://doi.org/10.5281/zenodo.3509134}.

\bibitem[Testolin et~al.(2020)Testolin, Piccolini, and
  Suweis]{testolin2020deep}
Testolin, A., Piccolini, M., and Suweis, S.
\newblock Deep learning systems as complex networks.
\newblock \emph{Journal Of Complex Networks}, 8\penalty0 (1):\penalty0 Cnz018,
  2020.

\bibitem[Tian et~al.(2019)Tian, Jiang, Gong, and Morcos]{tian2019luck}
Tian, Y., Jiang, T., Gong, Q., and Morcos, A.
\newblock Luck matters: Understanding training dynamics of deep relu networks.
\newblock \emph{ArXiv preprint}, abs/1905.13405, 2019.
\newblock URL \url{https://arxiv.org/abs/1905.13405}.

\bibitem[Tsai et~al.(2016)Tsai, Saxe, Saxe, and Cox]{NIPS2016_b1563a78}
Tsai, C.-Y., Saxe, A.~M., Saxe, A.~M., and Cox, D.
\newblock Tensor switching networks.
\newblock In Lee, D., Sugiyama, M., Luxburg, U., Guyon, I., and Garnett, R.
  (eds.), \emph{Advances in Neural Information Processing Systems}, volume~29.
  Curran Associates, Inc., 2016.
\newblock URL
  \url{https://proceedings.neurips.cc/paper/2016/file/b1563a78ec59337587f6ab6397699afc-Paper.pdf}.

\bibitem[Vaswani et~al.(2017)Vaswani, Shazeer, Parmar, Uszkoreit, Jones, Gomez,
  Kaiser, and Polosukhin]{transformer}
Vaswani, A., Shazeer, N., Parmar, N., Uszkoreit, J., Jones, L., Gomez, A.~N.,
  Kaiser, L., and Polosukhin, I.
\newblock Attention is all you need.
\newblock In Guyon, I., von Luxburg, U., Bengio, S., Wallach, H.~M., Fergus,
  R., Vishwanathan, S. V.~N., and Garnett, R. (eds.), \emph{Advances in Neural
  Information Processing Systems 30: Annual Conference on Neural Information
  Processing Systems 2017, December 4-9, 2017, Long Beach, CA, {USA}}, pp.\
  5998--6008, 2017.
\newblock URL
  \url{https://proceedings.neurips.cc/paper/2017/file/3f5ee243547dee91fbd053c1c4a845aa-Paper.pdf}.

\bibitem[Veness et~al.(2021)Veness, Lattimore, Bhoopchand, Budden, Mattern,
  Grabska-Barwinska, Toth, Schmitt, and Hutter]{Veness2021GatedLN}
Veness, J., Lattimore, T., Bhoopchand, A., Budden, D., Mattern, C.,
  Grabska-Barwinska, A., Toth, P., Schmitt, S., and Hutter, M.
\newblock Gated linear networks.
\newblock In \emph{AAAI}, 2021.

\bibitem[Vithayathil~Varghese \& Mahmoud(2020)Vithayathil~Varghese and
  Mahmoud]{electronics9091363}
Vithayathil~Varghese, N. and Mahmoud, Q.~H.
\newblock A survey of multi-task deep reinforcement learning.
\newblock \emph{Electronics}, 9\penalty0 (9), 2020.
\newblock ISSN 2079-9292.
\newblock \doi{10.3390/electronics9091363}.
\newblock URL \url{https://www.mdpi.com/2079-9292/9/9/1363}.

\bibitem[Wang et~al.(2019)Wang, Yu, Dunlap, Ma, Wang, Mirhoseini, Darrell, and
  Gonzalez]{wang2020deep}
Wang, X., Yu, F., Dunlap, L., Ma, Y., Wang, R., Mirhoseini, A., Darrell, T.,
  and Gonzalez, J.~E.
\newblock Deep mixture of experts via shallow embedding.
\newblock In Globerson, A. and Silva, R. (eds.), \emph{Proceedings of the
  Thirty-Fifth Conference on Uncertainty in Artificial Intelligence, {UAI}
  2019, Tel Aviv, Israel, July 22-25, 2019}, volume 115 of \emph{Proceedings of
  Machine Learning Research}, pp.\  552--562. {AUAI} Press, 2019.
\newblock URL \url{http://proceedings.mlr.press/v115/wang20d.html}.

\bibitem[Yadan(2019)]{Yadan2019Hydra}
Yadan, O.
\newblock Hydra - a framework for elegantly configuring complex applications.
\newblock Github, 2019.
\newblock URL \url{https://github.com/facebookresearch/hydra}.

\bibitem[Yang et~al.(2019)Yang, Bender, Le, and Ngiam]{yang2019condconv}
Yang, B., Bender, G., Le, Q.~V., and Ngiam, J.
\newblock Condconv: Conditionally parameterized convolutions for efficient
  inference.
\newblock In Wallach, H.~M., Larochelle, H., Beygelzimer, A.,
  d'Alch{\'{e}}{-}Buc, F., Fox, E.~B., and Garnett, R. (eds.), \emph{Advances
  in Neural Information Processing Systems 32: Annual Conference on Neural
  Information Processing Systems 2019, NeurIPS 2019, December 8-14, 2019,
  Vancouver, BC, Canada}, pp.\  1305--1316, 2019.
\newblock URL
  \url{https://proceedings.neurips.cc/paper/2019/file/f2201f5191c4e92cc5af043eebfd0946-Paper.pdf}.

\bibitem[Yang et~al.(2020)Yang, Xu, WU, and
  Wang]{multi-task-rl-with-soft-modularization}
Yang, R., Xu, H., WU, Y., and Wang, X.
\newblock Multi-task reinforcement learning with soft modularization.
\newblock In Larochelle, H., Ranzato, M., Hadsell, R., Balcan, M.~F., and Lin,
  H. (eds.), \emph{Advances in Neural Information Processing Systems},
  volume~33, pp.\  4767--4777. Curran Associates, Inc., 2020.
\newblock URL
  \url{https://proceedings.neurips.cc/paper/2020/file/32cfdce9631d8c7906e8e9d6e68b514b-Paper.pdf}.

\bibitem[Yuksel et~al.(2012)Yuksel, Wilson, and Gader]{yuksel2012twenty}
Yuksel, S.~E., Wilson, J.~N., and Gader, P.~D.
\newblock Twenty years of mixture of experts.
\newblock \emph{Ieee Transactions On Neural Networks And Learning Systems},
  23\penalty0 (8):\penalty0 1177--1193, 2012.

\bibitem[Zagoruyko \& Komodakis(2016)Zagoruyko and
  Komodakis]{zagoruyko2016wide}
Zagoruyko, S. and Komodakis, N.
\newblock Wide residual networks.
\newblock In Wilson, R.~C., Hancock, E.~R., and Smith, W. A.~P. (eds.),
  \emph{Proceedings of the British Machine Vision Conference 2016, {BMVC} 2016,
  York, UK, September 19-22, 2016}. {BMVA} Press, 2016.
\newblock URL \url{http://www.bmva.org/bmvc/2016/papers/paper087/index.html}.

\bibitem[Zambra et~al.(2020)Zambra, Maritan, and Testolin]{zambra2020emergence}
Zambra, M., Maritan, A., and Testolin, A.
\newblock Emergence of network motifs in deep neural networks.
\newblock \emph{Entropy}, 22\penalty0 (2):\penalty0 204, 2020.

\bibitem[Zhang et~al.(2014)Zhang, Luo, Loy, and
  Tang]{zhang2014facial_landmark_detection_by_deep_multitask_learning}
Zhang, Z., Luo, P., Loy, C.~C., and Tang, X.
\newblock Facial landmark detection by deep multi-task learning.
\newblock In \emph{European Conference On Computer Vision}, pp.\  94--108.
  Springer, 2014.

\end{thebibliography}
\bibliographystyle{icml2022}

\newpage
\appendix
\onecolumn

\section{Simple Nonlinear Classification}
\label{sec:nonlinear_contextual_classification}

\subsection{The XoR task}

We consider the XoR task with $P=4$ data points that lie at $[\pm1~\pm1]^T$ as depicted in Fig.~\ref{fig:xor}a. The task is exclusive-or on the input bits, with a target output of $y=1$ if true and $y=-1$ otherwise. We solve this with a GDLN containing four pathways (Fig.~\ref{fig:xor}b), with each pathway active on exactly one of the four examples. By symmetry, all pathways will have the same loss dynamics and so we need only solve one. Consider the pathway active for the example $x=[1~1]^T,y=1$, that is, whose gating variable $g_1=1$ when this example is presented and $g_1=0$ on the remaining three examples. The relevant dataset correlations are
\begin{eqnarray}
  \Sigma^{yx}&=&\langle g_1yx^T \rangle = 1/P\begin{bmatrix} 1 & 1\end{bmatrix} \\
  \Sigma^{x}&=&\langle g_1xx^T\rangle = 1/P\begin{bmatrix} 1 & 1 \\ 1 & 1\end{bmatrix} \\
  \Sigma^{y}&=&\langle g_1yy^T\rangle = 1/P
\end{eqnarray}

The singular value decomposition of the input-output correlations yields the nonlinear singular value $s=\sqrt{2}/P$ and input singular vector $v=[ 1/\sqrt{2} ~ 1/\sqrt{2}]^T$. Applying this singular vector to diagonalize the input correlations yields the associated input variance $d=v^T\Sigma^{x}v=2/P$. The effective singular value dynamics of this pathway is given by the deep linear network dynamics with these correlations (see \citet{Saxe2014,saxe_mathematical_2019}), yielding
\begin{equation}
    a(t)=\frac{s/d}{1-(1-\frac{s}{da_0})e^{-2st/\tau}}
\end{equation}
where $a(t)$ is the singular value in the product of both weight matrices in the pathway, and $a_0$ is the initial effective singular value, related to the initialization variance. Finally the loss for this pathway is the loss trajectory $l(t)$ of the associated deep linear network. The total loss, by symmetry, is the loss from all four pathways $\mathcal{L}(t)=Pl(t)$, 
\begin{eqnarray}
   \mathcal{L}(t) & = &\frac{1}{2} - Psa(t) + \frac{P}{2}da(t)^2 \\
   & = & \frac{1}{2} - \sqrt{2}a(t) + a(t)^2.
\end{eqnarray}
This analytical expression is exact for GDLNs initialized in the decoupled regime, and it agrees closely with the dynamics of standard ReLU networks trained end-to-end on the task starting from small random weights (Fig.~\ref{fig:xor}c). Hence GDLNs can learn nonlinear input-output tasks, and in certain settings, describe the dynamics of standard ReLU networks when the gating structure is chosen appropriately.

\subsection{Nonlinear Contextual Classification}
As another simple example, consider a \textbf{nonlinear contextual classification} problem that cannot be solved using deep linear networks but can be solved using the gated deep linear network, again highlighting that the gated networks are more expressive than their non-gated counterpart.

Consider receiving two-dimensional
inputs $x \in \R^2$ where each component $x_i, i=1,2$ is drawn from a uniform distribution between -1 and 1. The task of the network is to classify stimuli based either on the first or second input component. That is, the target output is $y=x_c$ in context $c\in {1,2}$, and each context appears with probability 1/2. In this simple scenario (a variant of the XoR task), the same input must be treated in two different ways depending on context, and nonlinearity is required for solving it correctly.

Now we must choose a gating structure. If we choose a single pathway that is
always active ($g=1$ for all samples), then we recover a deep linear network. The resulting
correlation matrices are
\begin{eqnarray}
  \Sigma^{yx}&=&\langle yx^T \rangle = [1/6~ 1/6] \\
  \Sigma^{x}&=&\langle xx^T\rangle = 1/3I
\end{eqnarray}
where $I$ is the $2\times 2$ identity matrix.
Under the resulting dynamics, the total weights converge to the linear least squares solution
$W^{tot}=\Sigma^{yx}(\Sigma^{x})^{-1}=[1/2 ~ 1/2]$,  the best solution attainable by the
linear network.

Alternatively, we can set the gating variables such that a different pathway
is active in each context. We then have the collection of correlation matrices
\begin{eqnarray}
  \Sigma^{yx}(1)&=& [1/6 ~0]\\
  \Sigma^{yx}(2)&=&[0 ~ 1/6]\\
  \Sigma^{x}(1,1)&=&\Sigma^{x}(2,2)=1/6I\\
  \Sigma^{x}(1,2)&=&\Sigma^{x}(2,1)=0.
\end{eqnarray}
We thus see that each pathway faces a subproblem defined by just one context. For this simple case,
the pathways converge to their respective linear least squares solutions. In particular,
$W^{tot}(1)=[1 ~ 0]$ and $W^{tot}(2)=[0 ~ 1]$, such that each pathway picks out the correct input
coordinate. In combination with the gating scheme, these weights exactly solve this nonlinear task, showing that gated linear networks are more expressive than linear networks.
Interestingly, neuroimaging and electrophysiological recordings from this paradigm suggest that this
type of solution is observed in the human and primate brain, as well as in standard ReLU networks
trained in the ``rich'' feature learning regime \cite{flesch_orthogonal_2022}.

\section{Gradient flow dynamics}
\label{apx:gradientflow}

The gradient flow equations are
\begin{eqnarray}
  \tau \frac{d}{dt}W_e & = & -\frac{\partial \mathcal L(\{W\})}{\partial W_e} \quad \forall e \in E \\
  & = & \left\langle \sum_{p\in \mathcal P(e)} g_pW_{\bar t(p,e)}^T\left[y_{t(p)}x^T_{s(p)} -  h_{t(p)}x_{s(p)}^T \right] W^T_{\bar s(p,e)} \right\rangle_{y,x,g}\\
  & = & \left\langle \sum_{p\in \mathcal P(e)} g_pW_{\bar t(p,e)}^T\left[y_{t(p)}x^T_{s(p)} - \sum_{j \in \mathcal T(t(p))} g_jW_jx_{s(j)}x_{s(p)}^T \right]W^T_{\bar s(p,e)} \right\rangle_{y,x,g}\\
  & = &  \sum_{p\in \mathcal P(e)} W_{\bar t(p,e)}^T\left[\left\langle g_py_{t(p)}x^T_{s(p)}\right\rangle_{y,x,g} - \sum_{j \in \mathcal T(t(p))} W_j\left\langle g_jx_{s(j)}x_{s(p)}^Tg_p \right\rangle_{y,x,g} \right]W^T_{\bar s(p,e)} \\
  & = &  \sum_{p\in \mathcal P(e)} W_{\bar t(p,e)}^T\left[\Sigma^{yx}(p) - \sum_{j \in \mathcal T(t(p))} W_j\Sigma^{x}(j,p) \right]W^T_{\bar s(p,e)}, \label{eq:weight_dyn}
\end{eqnarray}
where we have simply rearranged terms and used the linearity of expectation.

The dynamics reduction can then be obtained by applying the change of variables,
\begin{eqnarray}
  \tau \frac{d}{dt}W_e & = &  \sum_{p\in \mathcal P(e)} W_{\bar t(p,e)}^T\left[\Sigma^{yx}(p) - \sum_{j \in \mathcal T(t(p))} W_j\Sigma^{x}(j,p) \right]W^T_{\bar s(p,e)} \\
  \tau \frac{d}{dt}\left(R_{t(e)}B_eR_{s(e)}^T\right) & = &  \sum_{p\in \mathcal P(e)} \left(U_{t(p)}B_{\bar t(p,e)}R_{t(e)}^T\right)^T\left[U_{t(p)}S(p) V_{s(p)}^T - \right. \\
  &&\left. \sum_{j \in \mathcal T(t(p))} U_{t(j)}B_jV_{s(j)}^TV_{s(j)}D(j,p) V_{s(p)}^T \right]\left(R_{s(e)}B_{\bar s(p,e)}V_{s(p)}^T\right)^T \\
  \tau \frac{d}{dt}B_e & = &  \sum_{p\in \mathcal P(e)} B_{\bar t(p,e)}\left[S(p)  -  \sum_{j \in \mathcal T(t(p))} B_jD(j,p)  \right]B_{\bar s(p,e)}
\end{eqnarray}
where we have used the fact that $R_v^TR_v=I$ for all nodes, and the fact that
$U_{t(j)}=U_{t(p)}$ by definition of the set $\mathcal T(t(p))$. From this we
see that if the $B$ variables are initially diagonal they remain so under the dynamics. In this case, the dynamics decouple and each element along the diagonal of the $B$
matrices evolves independently of the rest.
\section{Routing task and network reduction}
\label{apx:routing_network_reduction}

To understand the dynamics in the pathway network, we first collect the relevant input statistics. We have
\begin{eqnarray}
  \Sigma^{yx}(p)&=& \left\langle g_p y_{t(p)} x_{s(p)}^T\right\rangle \\
  & = & \textrm{Pr}(g_p=1)USV^T\\
  & = & \frac{1}{KM}USV^T\\
  \Sigma^x(j,p)&=&\left\langle g_jx_{s(j)}x_{s(p)}^Tg_p\right\rangle\\
  &=& \begin{cases}
    \frac{1}{KM}VDV^T \quad \textrm{if}~j=p \\
    0 \quad \textrm{otherwise}
  \end{cases}
\end{eqnarray}
because there are $KM$ total trained paths from input to output and all pathways are gated off except for the active pathway.

Inserting these data statistics into \eqref{eq:decoulped_dyn}, and assuming that initial singular values are equal for all input domains and all output domains (a reasonable approximation when starting from small random weights), we can track only the variables $B_1,B_2,$ and $B_3$ encoding the singular values in the input, hidden, and output weights respectively.

Next, we note that the first and third layer weights are active in $K$ tasks (all tasks originating from a given input or terminating at a given output, respectively), while second layer weights are active in all $KM$ tasks. This yields the reduced dynamics
\begin{eqnarray}
  \tau \frac{d}{dt}B_1 &=& \frac{1}{M} B_3B_2\left[S-B_3B_2B_1D\right] \\
  \tau \frac{d}{dt}B_2 &=& B_3B_1\left[S-B_3B_2B_1D\right] \\
  \tau \frac{d}{dt}B_3 &=& \frac{1}{M}B_2B_1\left[S-B_3B_2B_1D\right].
\end{eqnarray}

If we consider `balanced' initial conditions where $B_1(0)=B_3(0)$, we have
\begin{eqnarray}
  \tau \frac{d}{dt}B_1 &=& \frac{1}{M} B_2B_1\left[S-B_2B_1^2D\right] \\
  \tau \frac{d}{dt}B_2 &=& B_1^2\left[S-B_2B_1^2D\right],
\end{eqnarray}
recovering Eqns.~\eqref{eqn:routing_net_reduct_eq1}-\eqref{eqn:routing_net_reduct_eq2} of the main text.

To estimate the ratio of singular values in the first layer to that in the second, we consider its time derivative and calculate the steady state. We have
\begin{eqnarray}
  \frac{d}{dt} B_2/B_1 & = & B_1\left[S-B_2B_1^2D\right] - \frac{1}{M} B_2^2/B_1\left[S-B_2B_1^2D\right]\\
  0 & = & B_1 - \frac{1}{M}B_2^2/B_1\\
  B_2 & = & \sqrt{M}B_1.
\end{eqnarray}
Hence if training continues for long times (such that the error term does not become zero), the shared portion of the pathway changes more by a factor $\sqrt{M}$ (and this ratio does not depend on $K$).

We can extend this analysis to the case where all input-output routes are trained but the gating structure sends only $P$ paths through each hidden weight matrix, as considered in Section~\ref{sec:neuralracered}. With this gating scheme, the reduction is
\begin{eqnarray}
  \tau \frac{d}{dt}B_1 &=& \frac{\sqrt{P}}{M^2} B_2B_1\left[S-B_2B_1^2D\right] \\
  \tau \frac{d}{dt}B_2 &=& \frac{P}{M^2} B_1^2\left[S-B_2B_1^2D\right],
\end{eqnarray}
and the singular value ratio is
\begin{eqnarray}
  \frac{d}{dt} B_2/B_1 & = & \frac{P}{M^2}B_1\left[S-B_2B_1^2D\right] - \frac{\sqrt{P}}{M^2} B_2^2/B_1\left[S-B_2B_1^2D\right]\\
  0 & = & PB_1 - \sqrt{P}B_2^2/B_1\\
  B_2 & = & P^{\frac{1}{4}}B_1.
\end{eqnarray}
This ratio scales from 1 to $\sqrt{M}$ as the number of shared paths goes from $P=1$ (no sharing) to $P=M^2$ (full sharing). Hence for this architecture, greater sharing causes larger weight changes in the hidden pathway.

\section{Experimental details and further results}
\label{apx:experiments}
\begin{figure*}
  \begin{center}
    \includegraphics[width=0.45\textwidth]{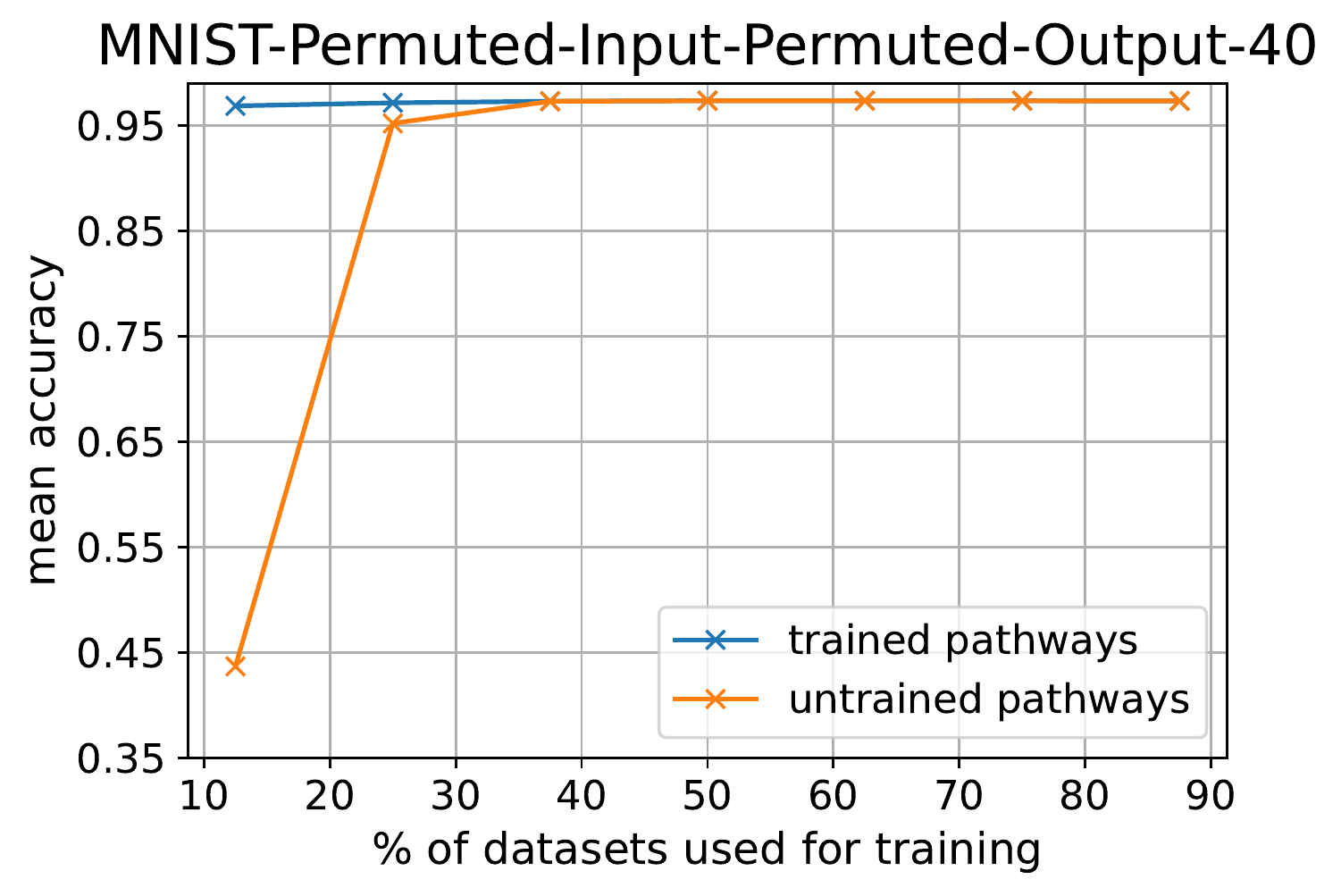}
    \includegraphics[width=0.45\textwidth]{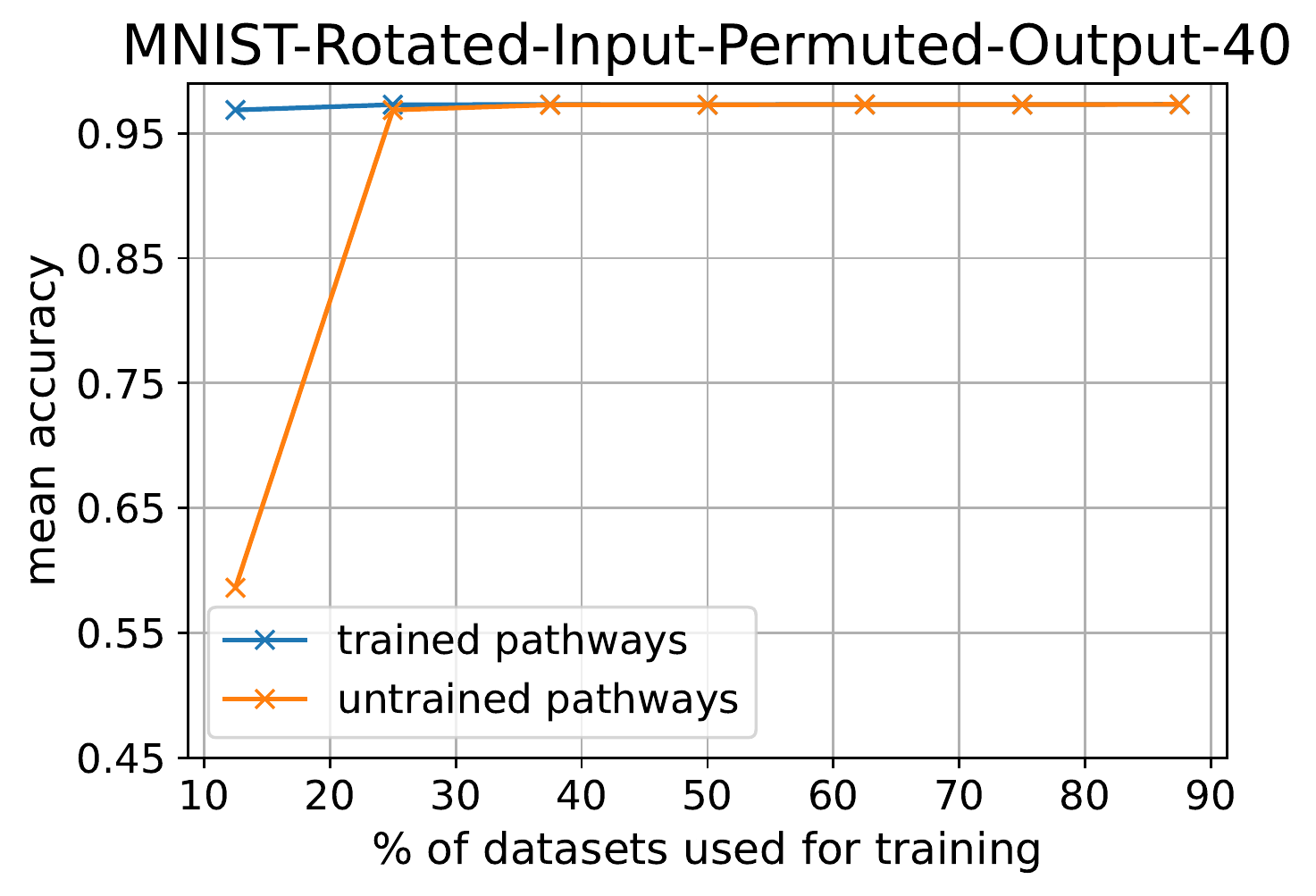}
  \end{center}

  \vspace{-3mm}
  \caption{Experimental results. Error for trained (blue) and untrained (orange) input-output domain pairs as a function of the percentage of trained pathways ($K/M$) on: (i) MNIST dataset, where input and output, both are permuted, with $M^2=1600$ total tasks, and (ii) MNIST dataset, where input is rotated and output is permuted, with $M^2=1600$ total tasks (in the order of left to right). These models are trained with the leaky-ReLU non-linearity with the negative slope parameter set to be $0.01$, thus making the model non-linear. The training accuracy is always high while zero-shot transfer to untrained pathways becomes as good as the training performance when $\approx$25\% of pathways are trained.}
\label{fig:results_on_natural_images2}
\end{figure*}

\begin{figure*}
  \begin{center}
    \includegraphics[width=0.3\textwidth]{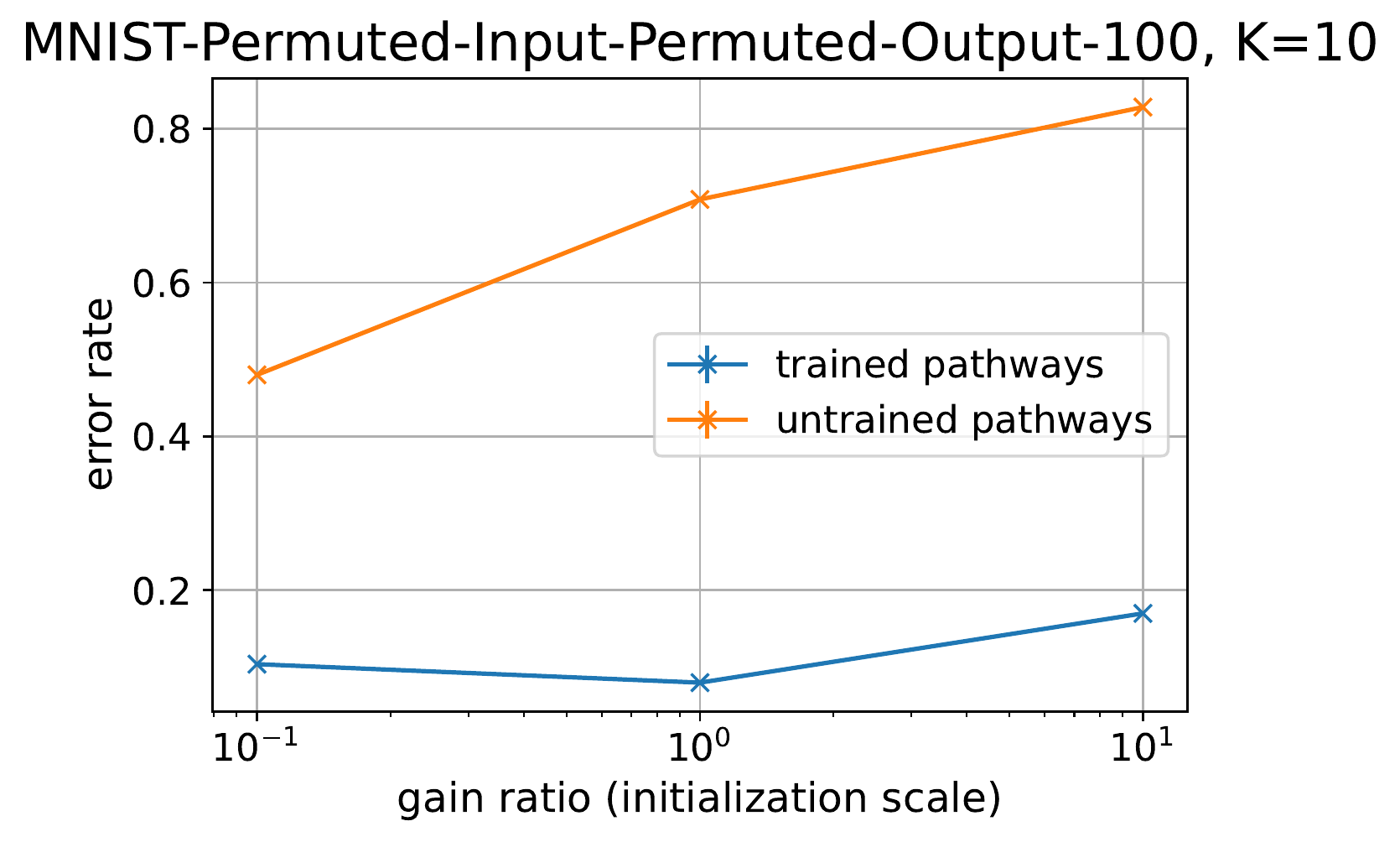}
    \includegraphics[width=0.3\textwidth]{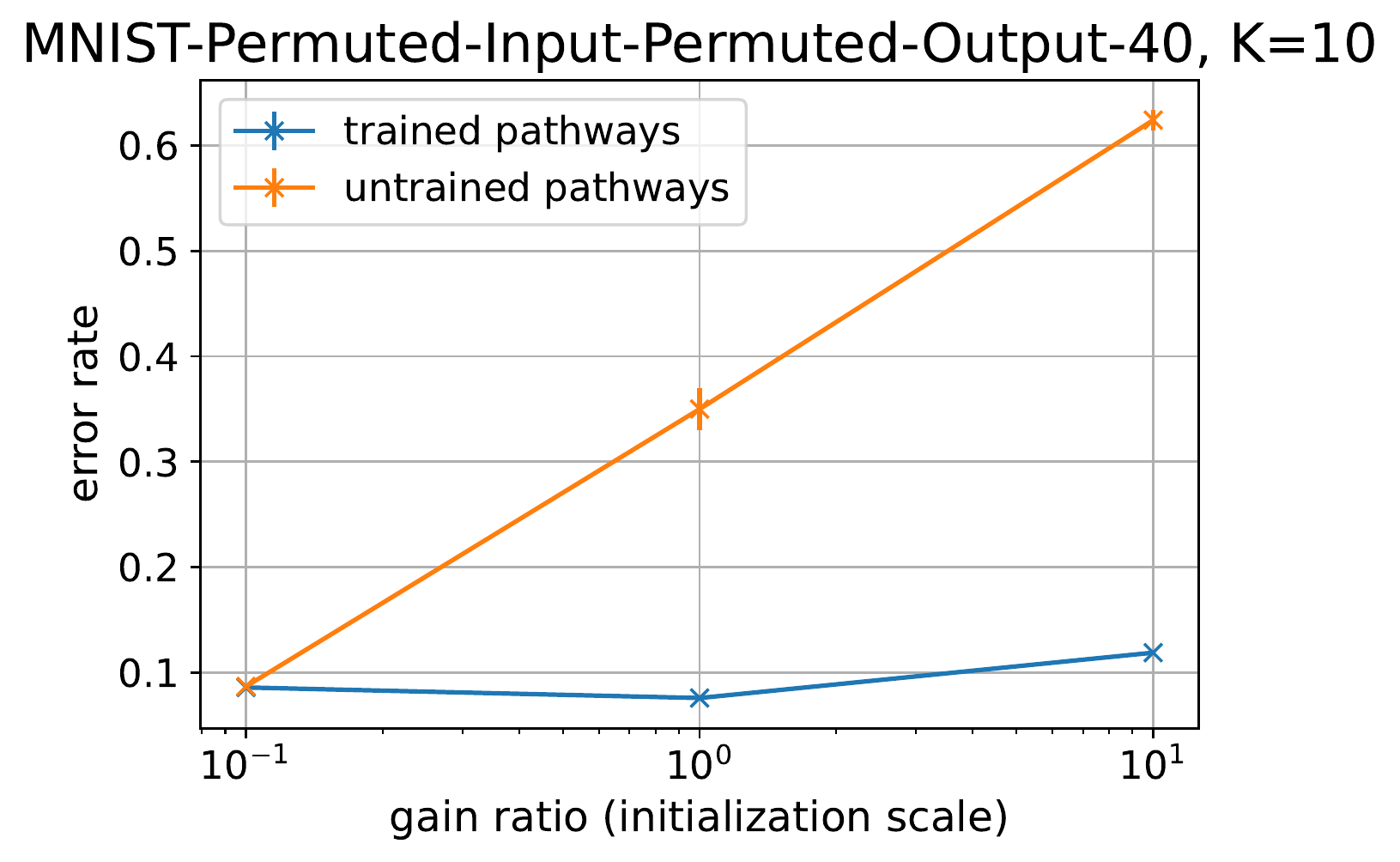}
    \includegraphics[width=0.3\textwidth]{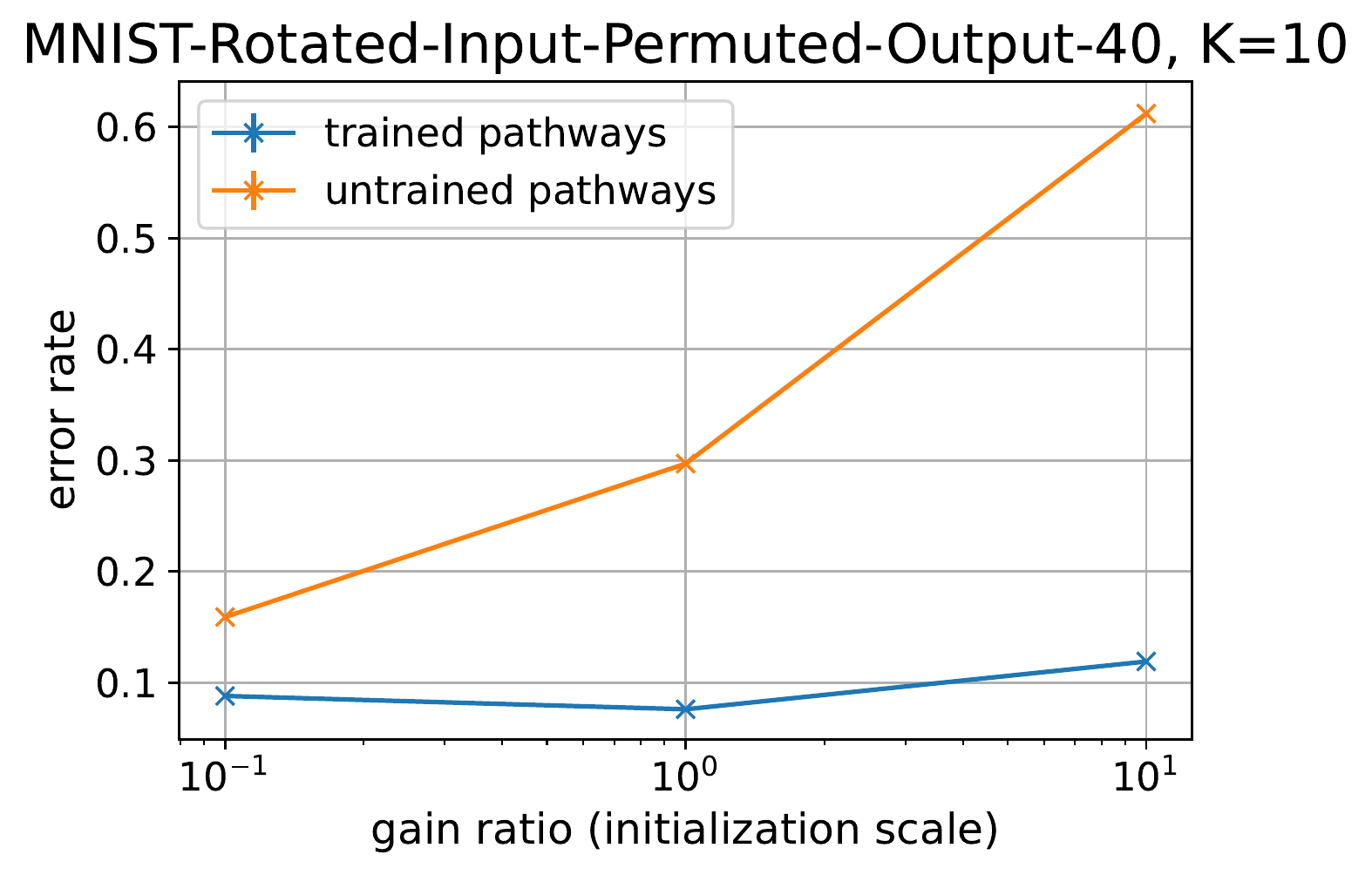}    
  \end{center}
\caption{Experimental results.  Error as a function of initialization scale for trained (blue) and untrained (orange) input-output domain pairs when training on $K/M$ fraction of pathways, where $K=10$ on (i) MNIST dataset, where input and output, both are permuted, with $M^2=10^4$ total tasks, (i) MNIST dataset, where input and output, both are permuted, with $M^2=1600$ total tasks, and (iii) MNIST dataset, where input is rotated and output is permuted, with $M^2=1600$ total tasks. (in the order of left to right). While performance on trained domains is reasonably good for all scales, zero-shot generalization only emerges at small inits.}

\section{Additional Implementation Details}
\label{app::implementation}

\subsection{Libraries}

We use the following open-source libraries:
\begin{enumerate}
    \item PyTorch~\cite{2019pytorch}\footnote{\url{https://pytorch.org/}}
    \item Hydra~\cite{Yadan2019Hydra}\footnote{\url{https://github.com/facebookresearch/hydra}}
    \item Numpy~\cite{harris2020array}\footnote{\url{https://numpy.org/}}
    \item Pandas~\cite{reback2020pandas}\footnote{\url{https://pandas.pydata.org/}}
    \item ResNet Implementation for CIFAR10/CIFAR100 in PyTorch~\cite{Idelbayev18a}\footnote{\url{https://github.com/akamaster/pytorch_resnet_cifar10}}
    \item Xplogger~\cite{Sodhani_xplogger_Logging_utility_2022}\footnote{\url{https://github.com/shagunsodhani/xplogger}}
\end{enumerate}

We ran all the experiments with 10 seeds and report the mean as well as the standard error across the seeds.

\label{fig:results_on_natural_images3}
\end{figure*}

This section contains details and hyperparameter settings for the simulations reported in Section~\ref{sec:experiment}, as well as additional visualization of results in Figures \ref{fig:results_on_natural_images2} and \ref{fig:results_on_natural_images3}.

We start by explaining the general procedure for transforming the existing datasets and then describe the new dataset instances that we create.

Consider a dataset $D=(X, Y)$, defined as a tuple of inputs $X$ and targets $Y$. The dataset $D$ has $n$ datapoints, that is, $X \in R^{n \times n_{dim}}$, where $n_{dim}$ is the dimensionality of each input\footnote{While images are multi-dimensional arrays, they can be represented as flattened 1-d arrays.}. The dataset has $n_{class}$ unique classes, referred by their indices $\{0, 1, \cdots, n_{class}-1\}$. We want to create new datasets by transforming the given dataset $D$. We assume that we have a list of $M$ input transformations $f_i^{input}: R^{n_{dim}} \to  R^{n_{dim}} \forall i \in \{0, \cdots, M-1\}$ and $M$ output transformations $f_j^{output}: R^{n_{class}} \to  R^{n_{class}} \forall j \in \{0, \cdots, M-1\}$. Now, we can define a new dataset, $D_{i,j}$, as $(X_i, Y_j)$, where $X_i = f_i^{input}(X)$, $Y_j = f_j^{output}(Y)$ i.e any transformation of the given dataset is a new dataset.
We can apply $M$ transformations on the input and $M$ transformations on the output to obtain $M^2$ datasets.

We consider the following two operations for input transformations:~\textit{rotation} of the input image and~\textit{permutation} of pixels. For the $i^{th}$~\textit{rotation} transformation, we rotate the input images by an angle $\theta = 180i/M$ degrees. For the $i^{th}$~\textit{permutation} transformation, we apply a random permutation matrix to the flattened input. Each of these transformations provides input to one input domain.  We use the~\textit{permutation} operation as the output transformation, implying that each new output transformation corresponds to a $n_{class}$-way classification task.

We use the MNIST~\citep{deng2012mnist} and CIFAR-10 datasets~\citep{krizhevsky2009learning} to create three datasets:
\begin{description}
  \item[\textit{MNIST-Permuted-Input-Permuted-Output-40}:]  $40$ transformations on both the input and the output, leading to a total of $1600$ datasets. The input transformation is permutation of pixels and the output transformation is permutation of the targets.
  \item[\textit{MNIST-Permuted-Input-Permuted-Output-100}:] $100$ permutation transformations on both the input and the output, leading to a total of $10^4$ datasets.
  \item[\textit{MNIST-Rotated-Input-Permuted-Output-40}] 40 rotation transformations on the input and 40 permutation transformations on the output.
  \item[\textit{CIFAR-Rotated-Input-Permuted-Output-40}] 40 rotation transformations on the input and 40 permutation transformations on the output.
\end{description}

In the case of~\textit{CIFAR-Rotated-Input-Permuted-Output-40} dataset, use a pre-trained ResNet18~\cite{resnet} model\footnote{We use the following code for pre-training the models: \url{https://github.com/akamaster/pytorch_resnet_cifar10}} to map the images into $512$ dimensional vectors. We pretrain the ResNet18 model on full CIFAR-10 dataset, freeze the pre-trained model and use the first two residual blocks to encode the images from the transformed datasets. The output of the (frozen) ResNet encoder is used as input to the gated network.

\subsection{Model and training}
\label{sec:experiment_model}

The model consists of $M$ encoders, denoted as ($\{\phi_i \forall i \in \{1, \cdots, M\}\}$) and $M$ decoders, denoted as ($\{\psi_i \forall i \in \{1, \cdots, M\}\}$) and a shared hidden layer $\theta$. In the GDLN framework, the encoders correspond to the~\textit{input nodes}, the decoders correspond to the ~\textit{output nodes} and the connection from an encoder, to the hidden layer, to the decoder corresponds to a~\textit{path}. The encoders, decoders and the shared hidden layer are all instantiated as linear networks. Given a dataset, $D_{i,j}$, or $(x_i, y_j)$, we compute the prediction using the following function: $\psi_j(\theta(\phi_i(x_i)))$~\footnote{Note that we overload the notation to represent the components and the computation using the same symbol.}

\subsection{Training and Evaluation Setup}
\label{sec:experiment_training_and_evaluation_setup}

Following the setup in~\cref{sec:applications}, we train a subset of input-output domains such that each input domain is trained with only $K \leq M$ output domains, resulting in $M \times K$~\textit{trained} pathways and $M \times (M-K)$~\textit{untrained} pathways. During evaluation, we report the performance on both the ~\textit{trained} pathways and the~\textit{untrained} pathways.

\begin{figure*}
  \begin{center}
    \includegraphics[width=0.23\textwidth]{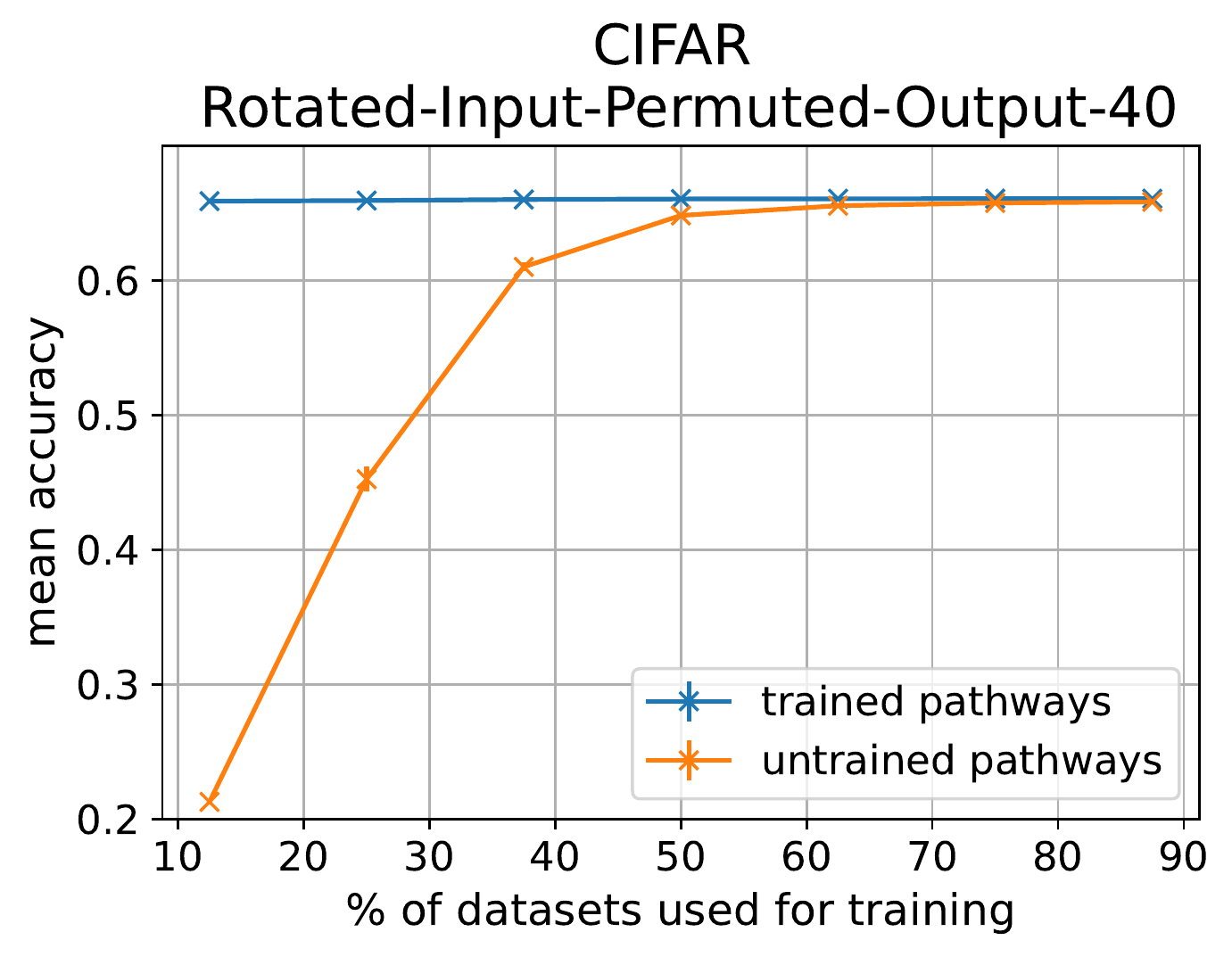}
    \includegraphics[width=0.23\textwidth]{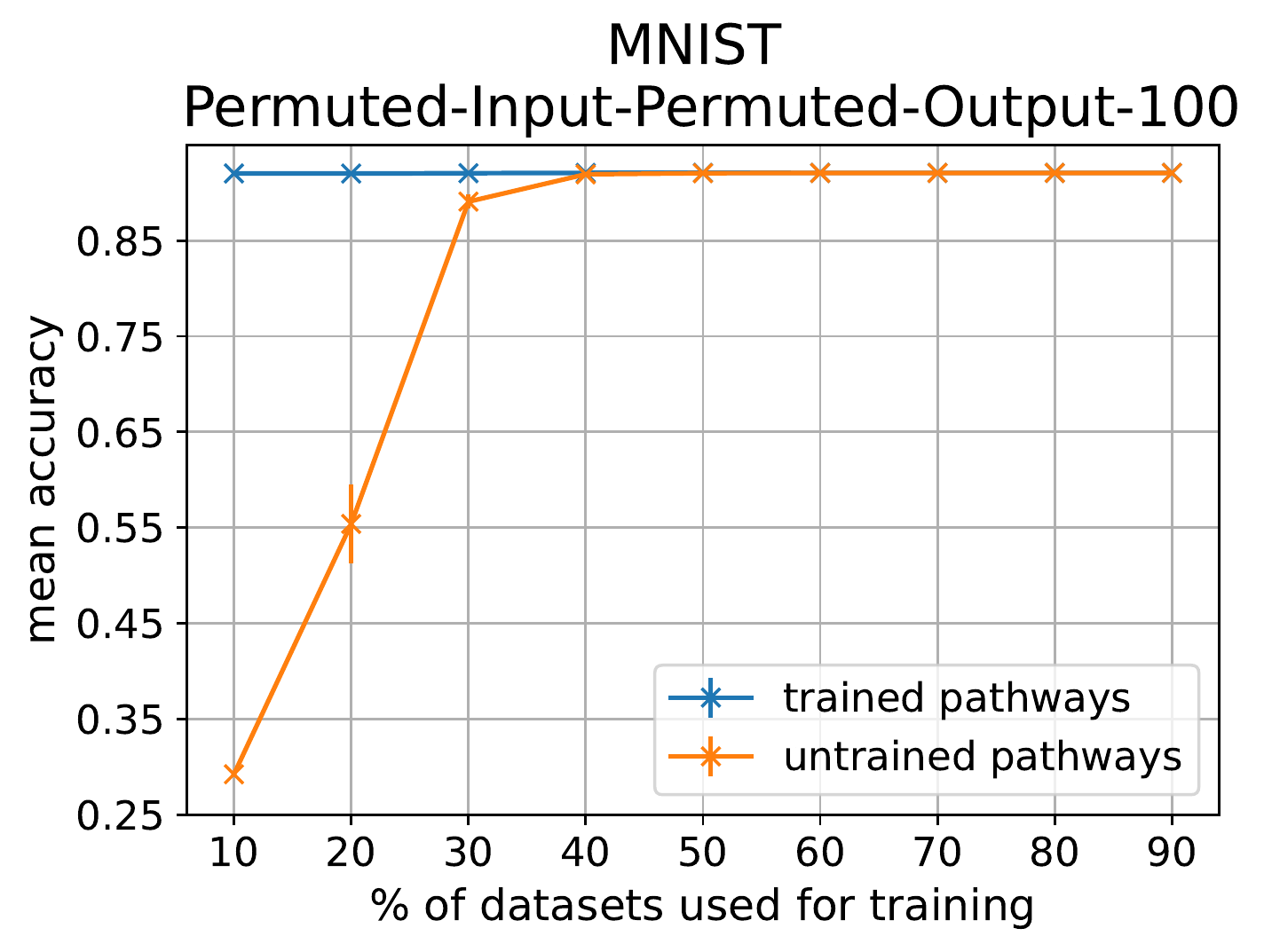}
    \includegraphics[width=0.23\textwidth]{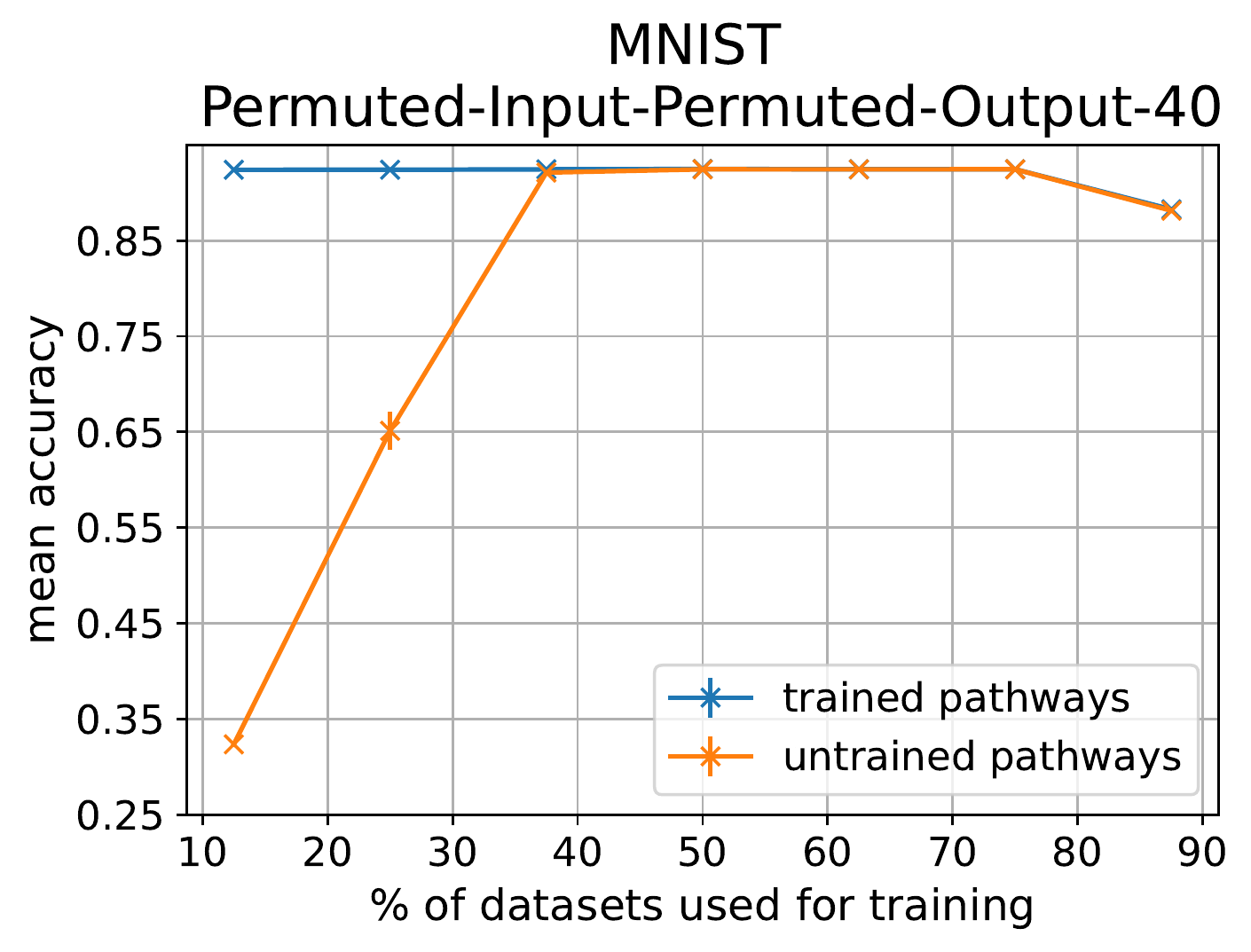}
    \includegraphics[width=0.23\textwidth]{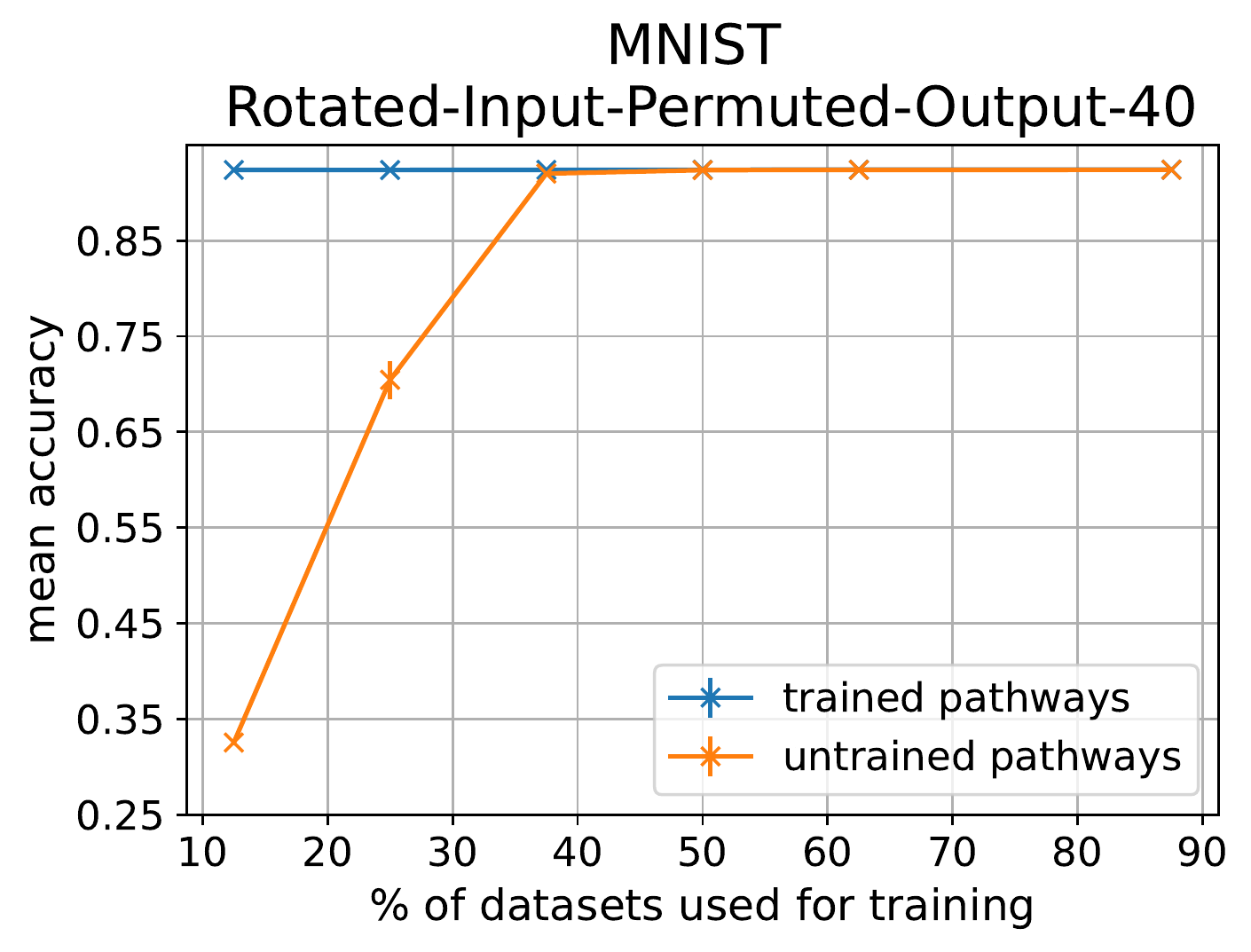}    
  \end{center}

  \vspace{-3mm}
  \caption{Experimental results. Error for trained (blue) and untrained (orange) input-output domain pairs as a function of the percentage of trained pathways ($K/M$) on: (i) CIFAR dataset, where input is rotated and output is permuted, with $M^2=1600$ total tasks, (ii) MNIST dataset, where input and output, both are permuted, with $M^2=10^4$ total tasks, (iii) MNIST dataset, where input and output, both are permuted, with $M^2=1600$ total tasks, and (iv) MNIST dataset, where input is rotated and output is permuted, with $M^2=1600$ total tasks. (in the order of left to right). Training accuracy is always high while zero-shot transfer to untrained pathways becomes as good as the training performance when $\approx$40\% of pathways are trained.}
\label{fig:results_on_natural_images2b}
\end{figure*}

\begin{table}[h]
  \centering
  \caption{Hyperparameter values common across all the task distributions}
  \label{table::common_hp}
  {
    \begin{tabular}{p{3.7cm}p{6.7cm}}
      \toprule
      Hyperparameter                     & Hyperparameter values            \\
      \midrule
      Batch size (per task)              & 8                                \\
      Number of epochs                   & $1000$                           \\
      Number of classes                  & $10$                             \\
      Dimensionality of the hidden layer & $128$                            \\
      Gain Ratio for weights             & $10.0,1.0,0.1,0.01,0.001,0.0001$ \\
      Initial value for bias             & $0.0$                            \\
      Optimizer                          & SGD                              \\
      Learning Rate                      & $0.0001$                         \\
      Momentum                           & $0.9$                            \\
      \bottomrule
    \end{tabular}
  }
\end{table}

\begin{table}[h]
  \centering
  \caption{Hyperparameter values for~\textit{MNIST-Permuted-Input-Permuted-Output-40} (different from the values described in~\cref{table::common_hp}}
  \label{table::hp_mnist_permuted_input_permuted_output_40}
  {
    \begin{tabular}{p{3.7cm}p{6.7cm}}
      \toprule
      Hyperparameter                   & Hyperparameter values \\
      \midrule
      Number of input transformations  & $40$                  \\
      Number of output transformations & $40$                  \\
      Number of encoders               & $40$                  \\
      Number of decoders               & $40$                  \\
      \bottomrule
    \end{tabular}
  }
\end{table}

\begin{table}
  \centering
  \caption{Hyperparameter values for~\textit{MNIST-Permuted-Input-Permuted-Output-100} (different from the values described in~\cref{table::common_hp}}
  \label{table::hp_mnist_permuted_input_permuted_output_100}
  {
    \begin{tabular}{p{3.7cm}p{6.7cm}}
      \toprule
      Hyperparameter                   & Hyperparameter values \\
      \midrule
      Number of input transformations  & $100$                 \\
      Number of output transformations & $100$                 \\
      Number of encoders               & $100$                  \\
      Number of decoders               & $100$                 \\
      \bottomrule
    \end{tabular}
  }
\end{table}

\begin{table}
  \centering
  \caption{Hyperparameter values for~\textit{MNIST-Rotated-Input-Permuted-Output-100} (different from the values described in~\cref{table::common_hp}}
  \label{table::hp_mnist_rotated_input_permuted_output_100}
  {
    \begin{tabular}{p{3.7cm}p{6.7cm}}
      \toprule
      Hyperparameter                   & Hyperparameter values \\
      \midrule
      Number of input transformations  & $100$                 \\
      Number of output transformations & $100$                 \\
      Number of encoders               & $100$                  \\
      Number of decoders               & $100$                 \\
      \bottomrule
    \end{tabular}
  }
\end{table}

\begin{table}
  \centering
  \caption{Hyperparameter values for~\textit{CIFAR-Rotated-Input-Permuted-Output-40} (different from the values described in~\cref{table::common_hp}}
  \label{table::hp_cifar_rotated_input_permuted_output_100}
  {
    \begin{tabular}{p{6.7cm}p{6.7cm}}
      \toprule
      Hyperparameter                   & Hyperparameter values \\
      \midrule
      Number of input transformations  & $40$                 \\
      Number of output transformations & $40$                 \\
      Number of encoders               & $40$                  \\
      Number of decoders               & $40$                 \\
      Dimensionality of the hidden layer & $1024$               \\
      \bottomrule
    \end{tabular}
  }
\end{table}

\end{document}